\documentclass[10pt,twocolumn,letterpaper]{article}

\usepackage{iccv}
\usepackage{times}
\usepackage{epsfig}
\usepackage{graphicx}
\usepackage{amsmath}
\usepackage{amssymb}
\usepackage{booktabs}

\usepackage[breaklinks=true,bookmarks=false]{hyperref}

\iccvfinalcopy %

\ificcvfinal\fi

\usepackage{algorithm}
\usepackage{listings}
\usepackage{etoolbox}
\makeatletter
\AfterEndEnvironment{algorithm}{\let\@algcomment\relax}
\AtEndEnvironment{algorithm}{\kern2pt\hrule\relax\vskip3pt\@algcomment}
\let\@algcomment\relax
\newcommand\algcomment[1]{\def\@algcomment{\footnotesize#1}}
\renewcommand\fs@ruled{\def\@fs@cfont{\bfseries}\let\@fs@capt\floatc@ruled
  \def\@fs@pre{\hrule height.8pt depth0pt \kern2pt}%
  \def\@fs@post{}%
  \def\@fs@mid{\kern2pt\hrule\kern2pt}%
  \let\@fs@iftopcapt\iftrue}
\makeatother
\usepackage{subcaption}

\usepackage{amsmath,amsfonts,bm}

\newcommand{\ba}[1]{\begin{align}#1\end{align}}

\def\Figref#1{Figure~\ref{#1}}

\def\eqref#1{equation~\ref{#1}}
\def\Eqref#1{Equation~\ref{#1}}

\def\1{\bm{1}}

\def\rvepsilon{{\boldsymbol{\epsilon}}}
\def\rvtheta{{\boldsymbol{\theta}}}
\def\rvphi{{\boldsymbol{\phi}}}

\def\rvc{{\boldsymbol{c}}}

\def\rvx{{\boldsymbol{x}}}

\def\vzero{{\bm{0}}}

\def\mI{{\bm{I}}}

\DeclareMathAlphabet{\mathsfit}{\encodingdefault}{\sfdefault}{m}{sl}
\SetMathAlphabet{\mathsfit}{bold}{\encodingdefault}{\sfdefault}{bx}{n}

\DeclareMathOperator*{\argmin}{arg\,min}

\newcommand{\x}[0]{\mathbf{x}}

\newcommand{\ldiff}[0]{\mathcal{L}_{\text{Diff}}}
\newcommand{\ldist}[0]{\mathcal{L}_{\text{SDS}}}

\usepackage{adjustbox}
\usepackage{graphicx}
\usepackage{dblfloatfix}
\usepackage{tikz-cd}

\begin{document}

\title{Re-imagine the Negative Prompt Algorithm:\\ Transform 2D Diffusion into 3D, alleviate Janus problem and Beyond}

\author{\vspace{0.2cm}Mohammadreza Armandpour$^{*1}$ \hspace{0.35cm} Ali Sadeghian$^{*2}$ \hspace{0.35cm} Huangjie Zheng$^{*3}$ \hspace{0.35cm} Amir Sadeghian$^2$ \hspace{0.35cm} Mingyuan Zhou$^3$ \\
$^1$Texas A\&M University \hspace{0.35cm} $^2$Astroblox AI \hspace{0.35cm} $^3$The University of Texas at Austin  \vspace{0.2cm} \\
\small\url{https://Perp-Neg.github.io/} \vspace{-0.8cm}
}

\maketitle
\ificcvfinal\thispagestyle{empty}\fi

\def\thefootnote{*}\footnotetext{Denotes equal contribution}

\begin{abstract}

Although text-to-image diffusion models have made significant strides in generating images from text, they are sometimes more inclined to generate images like the data on which the model was trained rather than the provided text. This limitation has hindered their usage in both 2D and 3D applications. To address this problem, we explored the use of negative prompts but found that the current implementation fails to produce desired results, particularly when there is an overlap between the main and negative prompts. To overcome this issue, we propose Perp-Neg, a new algorithm that leverages the geometrical properties of the score space to address the shortcomings of the current negative prompts algorithm. Perp-Neg does not require any training or fine-tuning of the model. Moreover, we experimentally demonstrate that Perp-Neg provides greater flexibility in generating images by enabling users to edit out unwanted concepts from the initially generated images in 2D cases. Furthermore, to extend the application of Perp-Neg to 3D, we conducted a thorough exploration of how Perp-Neg can be used in 2D to condition the diffusion model to generate desired views, rather than being biased toward the canonical views. Finally, we applied our 2D intuition to integrate Perp-Neg with the state-of-the-art text-to-3D (DreamFusion) method, effectively addressing its Janus (multi-head) problem.

\end{abstract}

\section{Introduction}

\begin{figure}
\centering
\includegraphics[width=0.8\columnwidth]{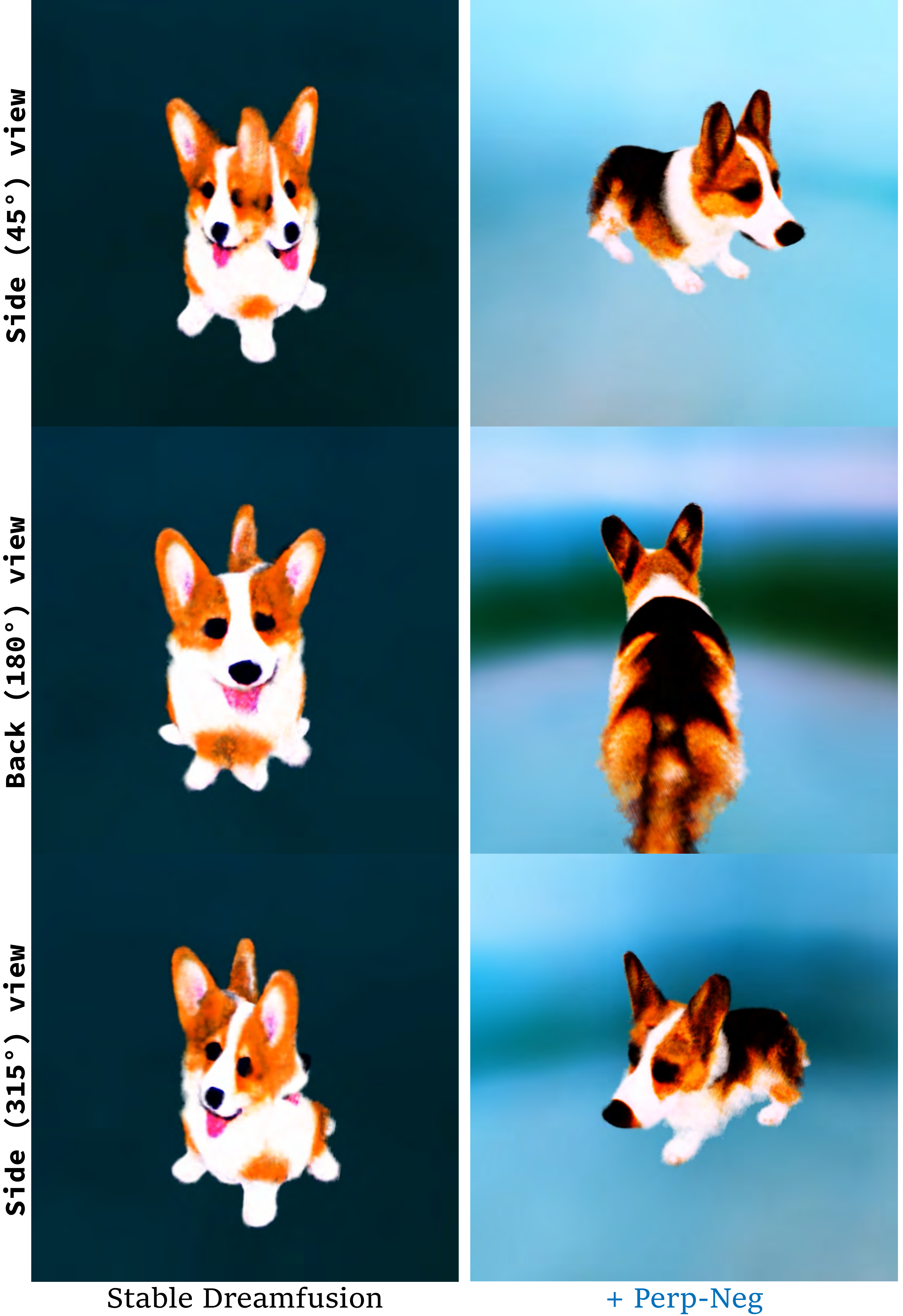}\vspace{-3mm}
\caption{Comparison of Stable-DreamFusion output with and without Perp-Neg algorithm for the prompt ``a corgi''. 
The Perp-Neg algorithm improves the accuracy of the 2D Diffusion model in following the view instructions specified in the text prompt during the training of the 3D scene. This helps to alleviate the Janus problem by encouraging the 2D diffusion to assign greater probability to the desired view in the text prompt instead of a canonical view.
}
\label{fig:janus}
\end{figure}

\begin{figure*}[t]
    \centering
    \includegraphics[width=0.8\textwidth]{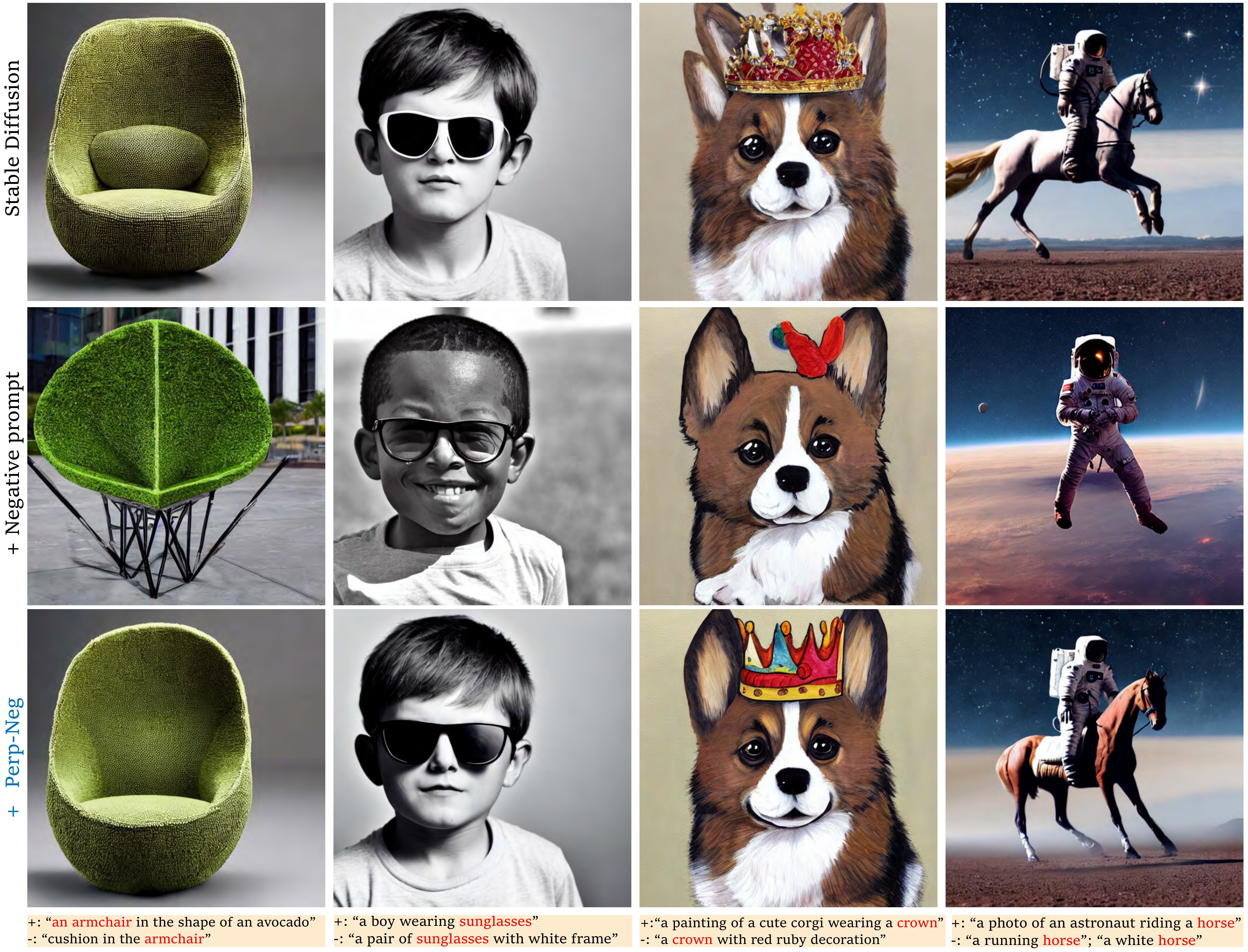}\vspace{-0.1mm}
    \caption{
    Illustration of Perp-Neg's (training-free) ability to modify generated images using negative prompts while preserving the main concept, for various combinations of positive (+) and negative (-) prompts.
    \textit{Top to Bottom:} Each column presents the generation from Stable Diffusion (using only positive prompt), Stable Diffusion using both positive and negative prompts, and Stable Diffusion with Perp-Neg sampling. The same seed has been used for the generation of each column.
    }
    \label{fig:motivation}\vspace{-2mm}
\end{figure*}

\begin{figure*}[h]
  \centering
  \includegraphics[width=0.8\textwidth]{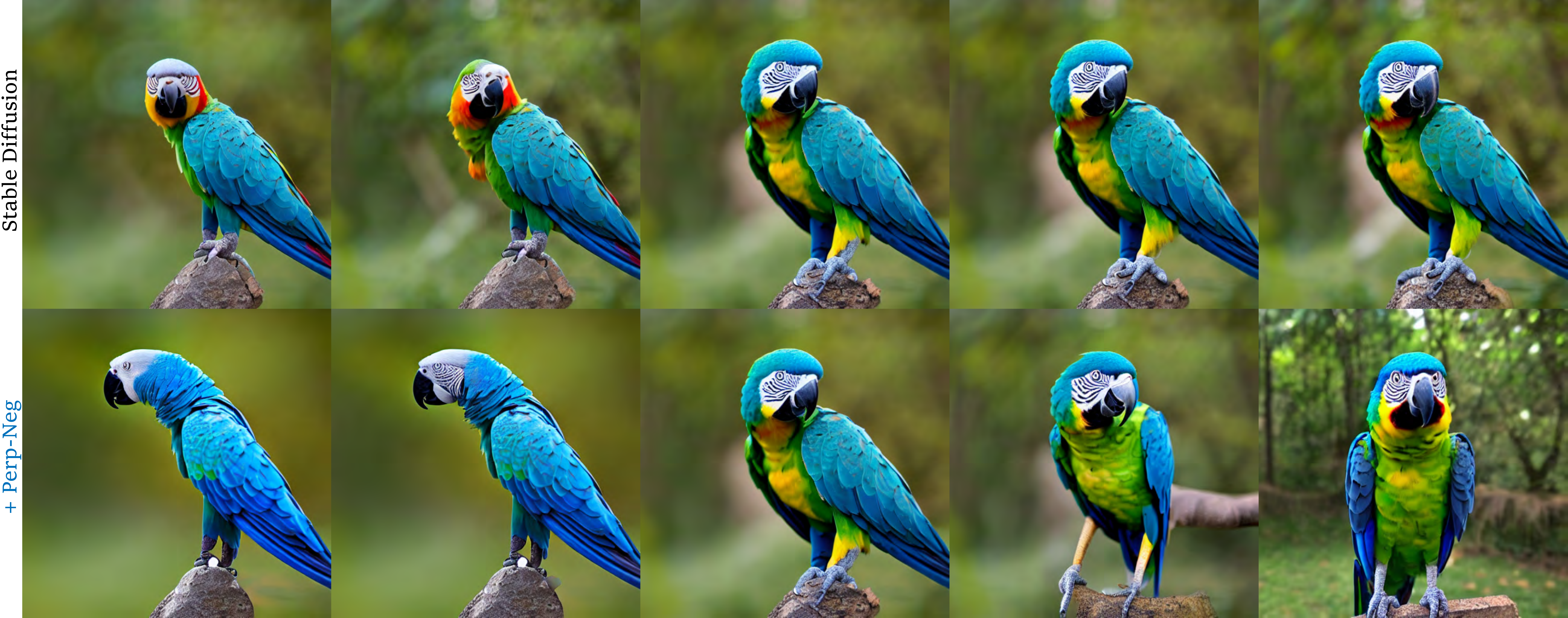} %
  \vspace{-10pt}
      \[
    \text{a photo of a parrot, side view} \xrightarrow{\quad\quad} \text{a photo of a parrot, front view}
    \]
  \vspace{-20pt}
  \caption{View interpolation with and without Perp-Neg. We fixed the seed across all different images.}
  \label{fig:parrot}
  \vspace{-10pt}
\end{figure*}

Advancements in generating images using diffusion models from text have shown remarkable capabilities in producing a wide range of creative images from unstructured text inputs~\cite{balaji2022ediffi, dalle2, rombach2022high,imagen,parti}. However, research has found that the generated images may not always accurately represent the intended meaning of the original text prompt~\cite{brooks2022instructpix2pix,chefer2023attend, hertz2022prompt, wangDiffusionDBLargescalePrompt2022}.

Generating satisfactory images that semantically match the text query is challenging, as it requires textual concepts to match the images at a grounded level. However, due to the difficulty of obtaining such a fine-grained annotation, current text-to-image models 
have difficulty fully understanding the relationship between text and images. Therefore, they are inclined to generate images like high-frequent text-image pairs in the datasets, 
where we can observe that the generated images are  missing requested or containing undesired attributes~\cite{li2023gligen}. 
Most of the recent works focus on adding back the missing objects or attributes to existing content to edit images based on a well-designed main text prompt~\cite{alt2022learning, brooks2022instructpix2pix,chefer2023attend, couairon2022diffedit,gal2022image,kawar2022imagic,lugmayr2022repaint, meng2021sdedit,su2022dual}. However, limited of them study how to remove redundant attributes, or force the model \textit{NOT} to have an unwanted object using negative prompts~\cite{du2020compositional}, which is the main goal of our paper. %

We start this paper by showing the shortcoming of the current negative prompt algorithm. %
After our initial investigation, we realized the current implementation of using negative prompts could produce unsatisfactory results when there is an overlap between the main prompt and the negative ones, as shown in the examples in \Figref{fig:motivation}. 
To address the above problem, we propose Perp-Neg algorithm, which does not require any training and can readily be applied to a pre-trained diffusion model. We refer to our method as Perp-Neg since it employs the perpendicular score estimated by the denoiser for the negative prompt. More specifically, Perp-Neg limits the direction of denoising, guided by the negative prompt to be always perpendicular to the direction of the main prompt. In this way, the model is able to eliminate the undesired perspectives in the negative prompts without changing the main semantics, as illustrated in \Figref{fig:motivation}.

Furthermore, We extend Perp-Neg to DreamFusion\cite{poole2022dreamfusion}, a state-of-the-art text-to-3D model,  and show how Perp-Neg can alleviate its Janus problem, which refers to the case that a 3D-generated object inaccurately shows the canonical view of the object from several viewpoints, as shown in the left column of \Figref{fig:janus}. Recent studies have considered that the main cause of the Janus problem is the failure of the pre-trained 2D diffusion model in following the view instruction provided in the prompt~\cite{metzer2022latent}.  Therefore, we first, in 2D, show quantitatively and qualitatively how our algorithm can significantly improve the view fidelity of a pretrained diffusion model. We also explore how Perp-Neg can be employed for effective interpolation between two views of an object in 2D as it is needed for 3D cases, as illustrated in Figure~\ref{fig:parrot}. Then we integrate Perp-Neg in Stable DreamFusion and show how it can alleviate the Janus problem.

Our contributions can be summarized as follows:
\begin{itemize}
    \item We find the limitations of the current negative prompt implementation which is susceptible to %
    the overlap between a positive and a negative prompt.
    \item We propose Perp-Neg, a sampling algorithm for text-to-image diffusion models to eliminate undesired attributes indicated by the negative prompt while preserving the main concept, without any training needed.     
    \item Our experiments quantitatively and qualitatively demonstrate that Perp-Neg significantly improves diffusion model prompt fidelity in view generation.%
    \item By enhancing the 2D diffusion model in following the view instruction, we mitigate the Janus problem in text-to-3D generation tasks. 
\end{itemize}

\section{Perp-Neg: Novel negative prompt algorithm}

\subsection{Preliminary} 
\textbf{Diffusion Models:} 
Diffusion-based (also known as score-matching) models~\cite{ddpm,sohl2015deep,song2021score} is a family of generative models that employ a forward process and a reverse process to iteratively corrupt and generate the data within $T$ steps. Specifically, denoting $q(\rvx_0)$ as the data distribution and $p(\rvx_T)$ as the generative prior, such two processes can be modeled as the following:
\ba{
&\text{forward}: q(\rvx_{0:T}) = q(\rvx_0) {\textstyle \prod_{t=1}^T} q(\rvx_t|\rvx_{t-1}), \notag \\
&\text{reverse}: p_\rvtheta(\rvx_{0:T}) = p(\rvx_T) {\textstyle \prod_{t=1}^T} p_\rvtheta(\rvx_{t-1}|\rvx_t).
}
One of the most appealing attributes of diffusion models is that any intermediate step of the forward process and every single step in the reverse process can be modeled as a Gaussian distribution like formulated in \cite{ddpm}:
\ba{
&q(\rvx_{t}|\rvx_{0}) = \mathcal{N}(\rvx_t; \sqrt{\alpha_t}\rvx_{0}, (1 - \alpha_t) \mI), \notag \\
&p_\rvtheta(\rvx_{t-1}|\rvx_t) = \mathcal{N}(\rvx_{t-1}; \mu_\rvtheta(\rvx_{t}, t), \sigma_t^2 \mI).
}
where $\{\alpha_t\}_{t=1}^T$ and $\{\sigma_t\}_{t=1}^T$ can be explicitly calculated with a pre-defined variance schedule $\{\beta_t\}_{t=1}^T$. Moreover, the generator $\mu_\rvtheta(\cdot)$ is a linear combination of $\rvx_t$ and a trainable generator $\rvepsilon_\rvtheta$ that predicts the noise in $\rvx_{t}$, which is usually optimized with a simple weighted noise prediction loss
\ba{
&\rvtheta^\star = \argmin_\rvtheta \mathop{\mathbb{E}}_{t, \rvx_t, \rvepsilon} [w(t) \| \rvepsilon_{\rvtheta}(\rvx_t,t) - \rvepsilon \|^2], \label{eq:noise_pred} \\
&\rvx_t = \sqrt{\alpha_t}\rvx_{0}+\sqrt{1-\alpha_t}\rvepsilon; ~\rvx_0 \sim q(\rvx_0), \rvepsilon \sim \mathcal{N}(\vzero, \mI) \notag
}
with $w(t)$ as the weight that depends on the timestep $t$ that is uniformly drawn from $\{1,...,T\}$.

\textbf{Text-to-Image Diffusion Models and Composing Diffusion Model:}        
Recent works have shown the success of leveraging the power of diffusion models, where large-scale models are able to be trained on extremely large text-image paired datasets by modeling with the loss function in \Eqref{eq:noise_pred} (or its variants)~\cite{glide,dalle2,rombach2022high,imagen}, with the text prompt $\rvc$ often encoded with a pre-trained large language model~\cite{bert,t5}. To generate photo-realistic images given text prompts, the diffusion models can further take advantage of classifier guidance \cite{dbeatgan} or classifier-free guidance \cite{ho2022classifier} to improve the image quality. Especially, in the context of text-to-image generation, classifier-free guidance is more widely used, which is usually expressed as a linear interpolation between the conditional and unconditional prediction 
$\hat \rvepsilon_\rvtheta(\rvx_t, t, \rvc) = (1+\tau)\rvepsilon_\rvtheta(\rvx_t, t, \rvc) - \tau \rvepsilon_\rvtheta(\rvx_t, t)$ at each timestep $t$ with a guidance scale parameter $\tau$.

When the prompt becomes complex, the model may fail to understand some key elements in the query prompt and create undesired images. To handle complex textual information, \cite{liu2022compositional} proposes composing diffusion models to factorize the text prompts into a set of text prompts, \textit{i.e.}, $\rvc \!\!=\!\! \{\rvc_1, ... \rvc_n \}$, and model the conditional distribution as
\ba{
\resizebox{0.90\hsize}{!}{%
        $%
p_{\rvtheta}(\rvx|\rvc_1, ..., \rvc_n)\propto p(\rvx, \rvc_1, ..., \rvc_n) = p_{\rvtheta}(\rvx) { \prod_{i=1}^n} {p_{\rvtheta}(\rvc_i|\rvx)}. $}\label{eq:composing_condition_distribution}
}
By applying Bayes rule, we have $p(\rvc_i|\rvx) \propto \frac{p(\rvx|\rvc_i)}{p(\rvx)}$ and 
\ba{
p_{\rvtheta}(\rvx|\rvc_1, ..., \rvc_n) \propto p_{\rvtheta}(\rvx) {\textstyle \prod_{i=1}^n} \frac{p_{\rvtheta}(\rvx|\rvc_i)}{p_{\rvtheta}(\rvx)}. \label{eq:composing_distribution}
} 
Note that $p_{\rvtheta}(\rvx|\rvc_i)$ and $p_{\rvtheta}(\rvx)$ respectively correspond to $\rvepsilon_\rvtheta(\rvx_t, t, \rvc_i)$ and $\rvepsilon_\rvtheta(\rvx_t, t)$ modeled by the diffusion model. Putting them together yields a composed noise predictor, as shown in \cite{liu2022compositional}: 
\ba{
\resizebox{0.90\hsize}{!}{%
        $
\hat \rvepsilon_\rvtheta(\rvx_t, t, \rvc) =\rvepsilon_\rvtheta(\rvx_t, t) + {\textstyle \sum_{i}} w_i \left( \rvepsilon_\rvtheta(\rvx_t, t, \rvc_i) - \rvepsilon_\rvtheta(\rvx_t, t) \right), \label{eq:composing_sampler}
$}
}
With $w_i$ as a scaling temperature parameter to adjust the weight of the concept components. When one concept $\tilde \rvc$ is needed to be removed, it is proposed to plug in the corresponding component $1/p(\rvx|\tilde \rvc)$ to reformulate \Eqref{eq:composing_distribution}:
\ba{
p_{\rvtheta}(\rvx|\text{not } \tilde \rvc, \rvc_1, ..., \rvc_n) = p_{\rvtheta}(\rvx) \frac{p_{\rvtheta}(\rvx)^\beta}{p_{\rvtheta}(\rvx|\tilde\rvc)^\beta} {\textstyle \prod_{i=1}^n} \frac{p_{\rvtheta}(\rvx|\rvc_i)}{p_{\rvtheta}(\rvx)}, \notag%
} 
and the corresponding sampler becomes
\ba{
\rvepsilon^\star_\rvtheta(\rvx_t, t, \rvc)\! =\! \hat \rvepsilon_\rvtheta(\rvx_t, t, \rvc)\! -\! w_\text{neg} \left(\rvepsilon_\rvtheta(\rvx_t, t, \tilde \rvc)\! -\! \rvepsilon_\rvtheta(\rvx_t, t)\right), \notag %
} 
where $w_\text{neg}>0$ is a weight function depending on $\tau$ and $\beta$, denoting the scale for the concept negation.

\subsection{Perpendicular gradient sampling}
\subsubsection{The problem of semantic overlap}
Although \cite{liu2022compositional} proposes to decompose the text condition into a set of positive and negative prompts in order to help the model handle complex textual inputs, the proposed method assumes these conditional prompts are independent of each other, which requires careful design of the prompts or maybe too ideal to realize in practice. 
For simplicity of presentation, below we present the overlap problem with the case of fusing two prompts, \textit{i.e.}, the main prompt $\rvc_1$ and an additional prompt $\rvc_2$. Without loss of generality, this problem can also be generalized to the case where the main prompt is combined with a series of prompts as $\{\rvc_1, ..., \rvc_n\}$.
To illustrate the problem, we first re-write the relation in \Eqref{eq:composing_condition_distribution}:
\ba{
p_{\rvtheta}(\rvx, \rvc_1, \rvc_2) \!=\! p_{\rvtheta}(\rvx) p_{\rvtheta}(\rvc_1| \rvx) p_{\rvtheta}(\rvc_2|\rvx) \frac{p_{\rvtheta}(\rvc_1, \rvc_2|\rvx)}{ p_{\rvtheta}(\rvc_1| \rvx) p_{\rvtheta}(\rvc_2|\rvx)}. \notag%
}
When $\rvc_1$ and $\rvc_2$ are conditional independent given $\rvx$, the ratio $\mathcal{R}(\rvc_1, \rvc_2) = \frac{p_{\rvtheta}(\rvc_1, \rvc_2|\rvx)}{ p_{\rvtheta}(\rvc_1| \rvx) p_{\rvtheta}(\rvc_2|\rvx)}\!=\!1$ and this term can be ignored. However, in practice, the input text prompts can barely be independent when we need to specify the desired attributes of the image, such as style, content, and their relations. When $\rvc_1$ and $\rvc_2$ have an overlap in their semantics, simply fusing the concepts could be harmful and result in undesired results, especially in the case of concept negation, as shown in \Figref{fig:motivation}. In the second row of images, we can clearly observe the key concepts requested in the main text prompt (respectively ``armchair", ``sunglasses", ``crown", and ``horse") are removed when those concepts appear in the negative prompts. This important observation motivates us to rethink the concept composing process and propose the use of a perpendicular gradient in the sampling, which is described in the following section.

\begin{figure*}[t]
    \centering
    \includegraphics[width=0.9\textwidth]{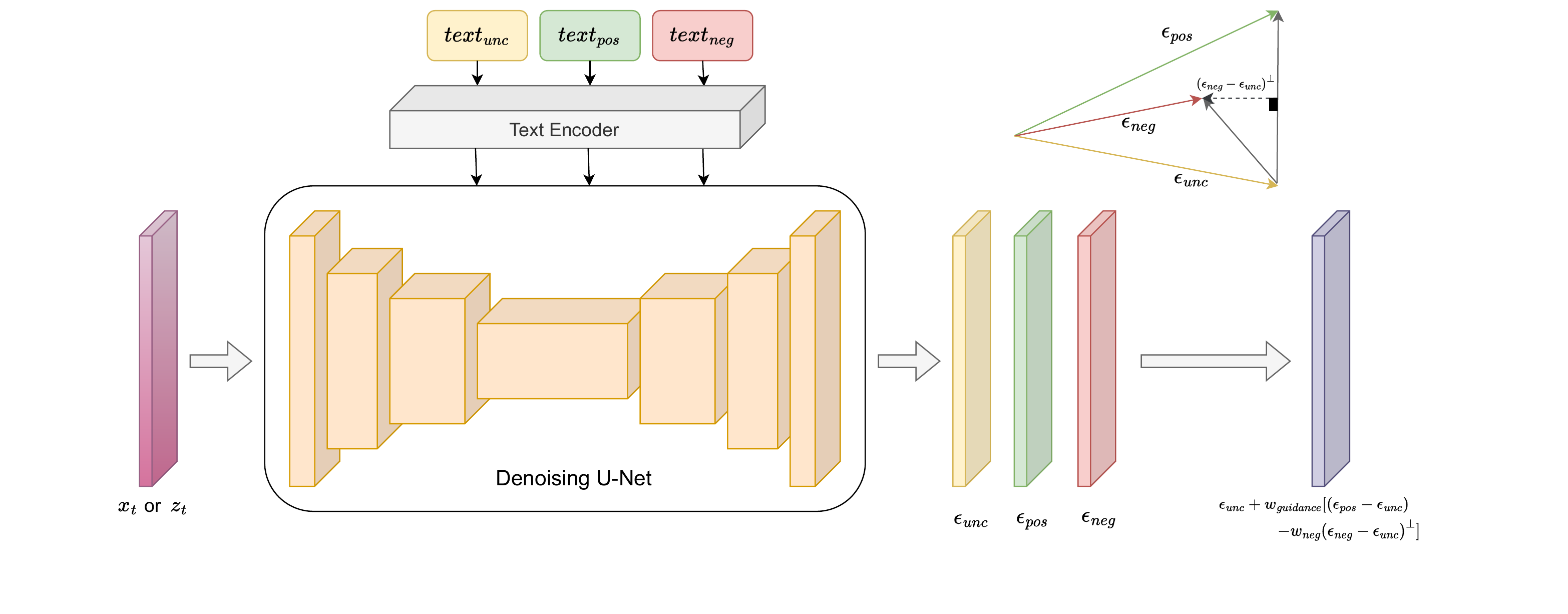}\vspace{-10mm}
    \caption{\textbf{Overview of Perp-Neg.} The plot shows a denoising step in the Perp-Neg algorithm for the whole scheme of 2D generation, refer to \Figref{fig:model} in Appendix}
    \label{fig:algo_diagram}\vspace{-5mm}
\end{figure*}

\subsubsection{Perpendicular gradient}
Recall when $\rvc_1$ and $\rvc_2$ are independent, both of them possess a denoising score component \vspace{-3mm}$$\rvepsilon_\rvtheta^i = \rvepsilon_\rvtheta(\rvx_t, t, \rvc_i) - \rvepsilon_\rvtheta(\rvx_t, t);~~i=1,2\vspace{-2mm}$$ and we can directly fuse these denoising scores as done in \Eqref{eq:composing_sampler}. However, from the above section, when $\rvc_1$ and $\rvc_2$ overlap, we cannot directly fuse the denoising components together, which motivates us to seek the independent component of $\rvc_2$ to ensure the fused denoising score does not hurt the semantics in $\rvc_1$.

Considering the geometrical interpretation of $\rvepsilon_\rvtheta^i$ indicates the gradient that the generative model should denoise to produce the final images, a natural solution is to find the perpendicular gradient of $\rvepsilon_\rvtheta^1$ as the independent component of $\rvepsilon_\rvtheta^2$. Therefore, we now re-formulate \Eqref{eq:composing_sampler} and define the Perp-Neg sampler for $\rvc_1$ and $\rvc_2$ as 
\ba{
\resizebox{0.90\hsize}{!}{%
        $
\hat \rvepsilon^\text{Perp}_\rvtheta(\rvx_t, t , \rvc) \!=\! \rvepsilon_\rvtheta(\rvx_t, t) \!+\! w_1 \rvepsilon_\rvtheta^1 \!+\! w_2 \!\!\underbrace{\left(\! \rvepsilon_\rvtheta^2 \!-\! \frac{\langle\rvepsilon_\rvtheta^1, \rvepsilon_\rvtheta^2\rangle}{\|\rvepsilon_\rvtheta^1\|^2} \rvepsilon_\rvtheta^1 \!\right)}_{\text{perpendicular gradient }}.
$}
\label{eq:perp_sampler}
} %
where $\langle, \rangle$ denotes the vectorial inner product, $w_1$ and $w_2$ define the weights for each component, and $\frac{\langle\rvepsilon_\rvtheta^1, \rvepsilon_\rvtheta^2\rangle}{\|\rvepsilon_\rvtheta^1\|^2}$ defines the projection function to find the most correlated component of $\rvc_2$ to $\rvc_1$. %

Note that although the proposed perpendicular gradient sampler is applicable for both positive text prompts and negative prompts, we find in the case of concept conjunction, the positive prompts can be designed to be independent of the main prompt in an easier way, as we are creating new details in complementary to the main concept. However, in the case of concept negation, it is more frequent to observe the negative prompts have overlap with the main text prompt. Compared to the sampler in \Eqref{eq:composing_sampler}, the most important property of the perpendicular gradient is that the component of $\rvepsilon_\rvtheta^1$ won't be affected by the additional prompt. Imagine the case where $\rvepsilon_\rvtheta^1=\rvepsilon_\rvtheta^2$, using \Eqref{eq:composing_sampler}, the denoising gradient becomes zero if we also set $w_1=-w_2$, which might fail the generation. However, using perpendicular gradient in \Eqref{eq:perp_sampler} could still preserve the main component $\rvepsilon_\rvtheta^1$. Below we mainly discuss the case of using perpendicular gradient sampling to handle the negative prompts and introduce Perp-Neg algorithm. 

\subsubsection{Perp-Neg algorithm}
The above section discusses the perpendicular gradient between the main prompt and one additional prompt. Here we generalize it to a set of negative text prompts $\{\tilde \rvc_1, ..., \tilde \rvc_m\}$ and present our Perp-Neg algorithm. We first denote $\rvc_1$ and $\rvepsilon_\rvtheta^1$ used in the previous section as $\rvc_\text{pos}$ and $\rvepsilon_\rvtheta^\text{pos}$, which indicate the main positive text prompt condition and the corresponding denoising component, respectively. For any negative text prompt in the set $\tilde \rvc_i,~i=1,...,m$, following \eqref{eq:perp_sampler}, the Perp-Neg sampler is defined as
\ba{
\rvepsilon_\rvtheta^\text{Perp-Neg}(\rvx_t, t, &\rvc_\text{pos},  \tilde \rvc_i) = \rvepsilon_\rvtheta(\rvx_t, t) + w_\text{pos}  \rvepsilon_\rvtheta^\text{pos} \notag \\
&- {\textstyle \sum_i}   w_i \!\!\!\!\!\underbrace{\left( \rvepsilon_\rvtheta^i - \frac{\langle\rvepsilon_\rvtheta^\text{pos}, \rvepsilon_\rvtheta^i\rangle}{\|\rvepsilon_\rvtheta^\text{pos}\|^2} \rvepsilon_\rvtheta^\text{pos}\right)}_{\text{perpendicular gradient of } \rvepsilon^\text{pos} \text{ on } \rvepsilon^i}\!\!\!\!,
}
with $\rvepsilon_\rvtheta^i = \rvepsilon_\rvtheta(\rvx_t, t, \tilde \rvc_i) - \rvepsilon_\rvtheta(\rvx_t, t)$, $w_\text{pos}>0$ and $w_i>0$ as the weight for positive and each negative prompt. The illustration of Perp-Neg algorithm is shown in \Figref{fig:algo_diagram}, and the detailed algorithm is described in Algorithm \ref{alg:code} in Appendix.

\section{2D diffusion model for 3D generation}
\textbf{Background:} Since 2D diffusion models not only provide samples of density but also allow calculating the derivate of data density likelihood. There are several seminal works that use the latter advantage to uplift a pretrained 2D diffusion and make it a 3D generative model. The main idea behind all these methods is to optimize a 3D scene representation of an object ($e.g.$, NeRF~\cite{mildenhall2021nerf}, mesh, $etc$.) based on the likelihood that a diffusion model defines its 2D projections. To be more specific, these algorithms consist of 3 main components:
\begin{itemize}
    \item 1- A 3D parametrization of the scene $\rvphi$. 
    \item 2- A differentiable renderer $g$ to create an image $\rvx$ (or its encoded feature) from a desired camera viewpoint $v$ such that  $\rvx = g(\rvphi, v)$. 
    \item 3- A pre-trained 2D diffusion model $\rvtheta$ to obtain a proxy of $\log p(\rvx |\rvc, v)$ where $p$ is the 2D data density and $\rvc$ is the text prompt.
\end{itemize}

The 3D generation has been done as solving an optimization problem as follows:
$$\rvphi^* = \argmin_{\rvphi} \mathbb{E}_{v}\left[\mathcal{L}(\rvx=g(\rvphi, v)| \rvc, v; \rvtheta)\right]$$
where $\mathcal{L}$ is a proxy to the negative log-likelihood of the 2D image based on the pre-trained diffusion model.

Remind the noise prediction loss in \Eqref{eq:noise_pred} is a natural choice for $\mathcal{L}$ as the training objective of the diffusion model, since it is a (weighted) evidence lower bound (ELBO) of the data density \cite{ddpm,kingma2021on, poole2022dreamfusion}:
\begin{align}
\ldiff =  \mathbb{E}_{t, \rvepsilon} & \left[w(t)\|\rvepsilon_\rvtheta(\rvx_t; t) - \rvepsilon\|^2_2\right] \, 
\label{eq:train}
\end{align}

However, direct optimization of $\ldiff$ does not provide realistic samples \cite{poole2022dreamfusion}. Therefore, Score Distillation Sampling (SDS) has been proposed as a modified version of the diffusion loss gradient $\nabla_{\rvphi}\ldiff$,  which is more robust and more computationally efficient as follows:
\begin{equation}\label{eq:sds_loss}
   \nabla_{\rvphi} \ldist(\rvx=g(\rvphi)) \triangleq \mathbb{E}_{t, \rvepsilon}\left[w(t)\left(\hat\rvepsilon_\rvtheta(\rvx_t; \rvc, v, t)  - \rvepsilon\right) {\partial \rvx \over \partial \rvphi}\right]
\end{equation}
where also $\rvepsilon_\rvtheta$ has been replaced with $\hat\rvepsilon_\rvtheta$ to allow text conditioning by using the classifier-free guidance~\cite{ho2022classifier}.

Intuitively, this loss perturbs $\rvx$ with a random amount of noise corresponding to the timestep $t$, and estimates an update direction that follows the score function of the diffusion model to move to a higher-density region.

For the choice of $\mathcal{L}$, since the introduction of the seminal work DreamFusion \cite{poole2022dreamfusion}, there have been several proposals \cite{lin2022magic3d, metzer2022latent, wang2022score}. However, since they are similar in core and our method can be applied to all of them, we continue the formulation of the paper by using the Score Distillation Sampling loss presented by DreamFusion.

\subsection{The Janus problem} 
Since the introduction of 2D diffusion-based 3D generative models, it has been known that they suffer from the Janus (multi-faced) problem \cite{metzer2022latent, poole2022dreamfusion}. This refers to a phenomenon that the learned 3D scene, instead of presenting the 3D desired output, shows multiple canonical views of an object in different directions. For instance, when the model is asked to generate a 3D sample of a person/animal, the generated object model has multiple faces of the person/animal (which is their canonical view) instead of having their back view. 

View-dependent prompting ($e.g.$, adding back view, side view, or overhead view with respect to the camera position to the main prompt) has been proposed as a remedy but does not fully solve the problem \cite{dreamfusion}. We believe part of the reason is that 2D Diffusion models fail to be fully conditioned on the view provided by the prompt, as also pointed out by others \cite{metzer2022latent}. For instance, when the model is asked to generate the back view of a peacock, it wrongly produces the front view instead, as the front view has been more prominent in the training data the model has been trained on.

To provide an intuitive mathematical understanding of the Janus problem, we believe one of the reasons is the model fails to  be properly conditioned on view $v$. More specifically, the proxy of $\log p(\rvx | \rvc, v)$ does not fully restrict $\rvx$ to have zero density on areas that do not represent the viewpoint $v$ for the scene description $y$. The main reason we think this is the case is samples of the density fail to reflect the direction of interest.

\subsection{Perp-Neg to alleviate Janus problem and 2D view conditioning}\label{}

In this section, we first explain how combining Perp-Neg with a unique prompting technique can enable us to accurately condition the 2D diffusion model on the desired view. Additionally, we will explore how Perp-Neg can be integrated with DreamFusion to address the Janus problem by improving the view faithfulness of the 2D model.

To begin, we demonstrate how to generate a desired statistical view using the improved model. Then, we explain the process for creating interpolations between two views
To generate a specific view of an object, we use a combination of positive and negative prompts. We define $\textbf{txt}_{\textit{back}}$, $\textbf{txt}_{\textit{side}}$, and $\textbf{txt}_{\textit{front}}$ as the main text prompts appended by back, side, and front views, respectively. We replace simple prompts containing the view with the following set of positive and negative prompts to generate each view:
\ba{
&\textbf{txt}_{\textit{back}}\rightarrow [+ \textbf{txt}_{\textit{back}},  
- w^\mathrm{b}_{\textit{side}} \textbf{txt}_{\textit{side}},  - w^\mathrm{b}_{\textit{front}} \textbf{txt}_{\textit{front}}] \notag\\
&\textbf{txt}_{\textit{side}}\rightarrow [+ \textbf{txt}_{\textit{side}},  - w^\mathrm{s}_{\textit{front}} \textbf{txt}_{\textit{front}}] \notag\\
&\textbf{txt}_{\textit{front}}\rightarrow [+ \textbf{txt}_{\textit{front}},  - w^\mathrm{f}_{\textit{side}} \textbf{txt}_{\textit{side}}] \notag
}
where $w_{(\cdot)} \geq 0$ denotes the weights for the negative prompts. Positive and negative prompts are fed into the Perp-Neg algorithm during each iteration of the diffusion model. We don't include $\textbf{txt}_\textit{back}$ as a negative prompt for the generation of side/front views since most objects' canonical view is not back. However, if the back view is more prominent for some objects, it should be included as a negative prompt. We also observed increasing the weight of the negative prompt makes the algorithm focus more on avoiding that view, acting as a pose factor.

In this subsection, we will first explain how we interpolate between the side and back views, followed by the interpolation between the front and side views. We distinguish between these two cases because the diffusion model may be biased toward generating front views, and if this assumption is not true, then the formulation needs to be adjusted accordingly.

To interpolate between the side and back views, we use the following embedding as the positive prompt:
$$ r_{\textit{inter}} * \textbf{emb}_{\textit{side}} + (1 - r_{\textit{inter}})\textbf{emb}_{\textit{back}}; \quad \quad 0 \leq r_{\textit{inter}} \leq 1 $$
where $\textbf{emb}_v$ is the encoded text for the view $v$ and $r_{\textit{inter}}$ is the degree of interpolation. And for the negative prompts, we use:
$$[-f_\mathrm{sb}(r_{\textit{inter}})\textbf{txt}_{\textit{side}},  - f_\mathrm{fsb}(r_{\textit{inter}}) \textbf{txt}_{\textit{front}}]$$
such that $f_\mathrm{sb}, f_\mathrm{fsb}$
are positive decreasing functions. 
The second negative prompt is chosen based on the assumption that the diffusion model is more biased towards generating samples from the front view.

For interpolation between the front and side views, the embedding for the positive would be:
$$ r_{\textit{inter}} * \textbf{emb}_{\textit{front}} + (1 - r_{\textit{inter}})\textbf{emb}_{\textit{side}}$$ and the following two negative prompts
$$[-f_\mathrm{fs}(r_{\textit{inter}})\textbf{txt}_{\textit{front}},  - f_\mathrm{sf}(1-r_{\textit{inter}}) \textbf{txt}_{\textit{side}}]$$
where $f_\mathrm{fs}(1), f_\mathrm{sf}(1) \approx 0$ and both of the functions are decreasing.

\textbf{Perp-Neg SDS:} We employed interpolation technique in Stable DreamFusion and varied $r_{\textit{inter}}$ based on the related direction of 3D to 2D rendering. To be more specific, we modified the SDS loss \ref{eq:sds_loss} as follows:
\begin{equation}\label{eq:sds_loss_perpneg}
   \nabla_{\rvphi} \ldist^{\textit{PN}} \triangleq \mathbb{E}_{t, \rvepsilon}\left[w(t)\left(\hat\rvepsilon^{\textit{PN}} _\rvtheta(\x_t; \rvc, v, t)  - \rvepsilon\right) {\partial \x \over \partial \rvphi}\right]
\end{equation}
such that $\hat\rvepsilon^{\textit{PN}} _\rvtheta(\rvx_t; \rvc, v, t)$ is:
\ba{
 \rvepsilon^\text{unc}_{\rvtheta} +w_{\textit{guidance}} [{\rvepsilon^{\text{pos}_v}_{\rvtheta}  - \sum_i w^{(i)}_v \rvepsilon^{\text{neg}^{(i)\perp}_{v}}_{\rvtheta} ].
}}
The unconditional term $\rvepsilon^{unc}_{\rvtheta}$ refers to $ \rvepsilon_\rvtheta(\rvx_t, t)$, and
\ba{
&\rvepsilon^{\text{pos}_v}_{\rvtheta} =\rvepsilon_\rvtheta(\rvx_t, t, c_\text{pos}^{(v)}) - \rvepsilon_\rvtheta(\rvx_t, t) \notag \\
&\rvepsilon^{\text{neg}^{(i)}_{v}}_{\rvtheta} =\rvepsilon_\rvtheta(\rvx_t, t, c_{\text{neg}_{(i)}}^{(v)}) - \rvepsilon_\rvtheta(\rvx_t, t) \notag
}
where $c_{.}^{(v)}$ refers to the text embedding of positive/negative at direction $v$. And $\rvepsilon^{\text{neg}^{(i)\perp}_{v}}_{\rvtheta}$ is the perpendical component of $\rvepsilon^{\text{neg}^{(i)}_{v}}_{\rvtheta}$ on $\rvepsilon^{\text{pos}_v}_{\rvtheta}$. And, $w_v$'s are representative of the weights of the negative prompts at direction $v$. 

\textbf{Further extension:} Although we only provide an application of Perp-Neg using SDS loss, we will investigate its application in  novel view synthesis~\cite{deng2022nerdi, liu2023zero, tang2023make, vinod2023teglo, watson2022novel}, conditional 3D generation~\cite{ Karnewar2023HOLODIFFUSION, nichol2022point, po2023compositional, raj2023dreambooth3d}, editing \cite{ceylan2023pix2video, haque2023instruct, mikaeili2023sked, voynov2023p+} and adding texture \cite{richardson2023texture}.

\section{Experiments}
In this section, we first conduct experiments on 2D cases to quantitatively demonstrate the importance of using Perp-Neg in the sampling to improve the likelihood of getting the image corresponding to the text query, which provides evidence of why our method surpasses vanilla sampling in the 3D case. Next, we show results in 3D generation.

\subsection{Statistics on semantic-aligned 2D generations}\label{sec:quantitative exp}
To understand why Perp-Neg improves the 3D generation quality, we first explore the 2D generation of the requested view to see whether Perp-Neg produces images with fewer artifacts than the vanilla sampling method. 

In the first experiment, we fix the random seeds as 0-49 to get 50 images from each text prompt. We carefully select qualified images that align with the requested text based on a series of criteria and report the percentage of accepted samples produced with Stable Diffusion, Compositional Energy-based Model (CEBM), and our Perp-Neg. 
Below we introduce the details of prompt design and the criteria for accepting qualified samples.

\noindent\textbf{Design of prompts:} We design the basic text prompts as: ``A [O], [V] view." Token [O] stands for the objects, such as panda, lion; token [V] stands for view, where we only consider ``front", ``back" and ``side" in our experiments. For example, we use ``A panda, side view" to request the model to generate an image showing the side view of a panda. 
We aim to test two groups of text prompts that generate the side view and the back view of the objects, which are considered simple and complex cases, respectively. For each group, when using the negative prompts, we use the complementary view or the combination of the other two views, \textit{e.g.}, in the case of using the ``side" view in the positive prompt, we use the ``front" view in the negative prompt. Both positive and negative prompts follow the basic prompt pattern but respectively adopt positive and negative weight in the fusing stage. 
\begin{figure}[t]
    \centering
    \includegraphics[width=0.9\columnwidth]{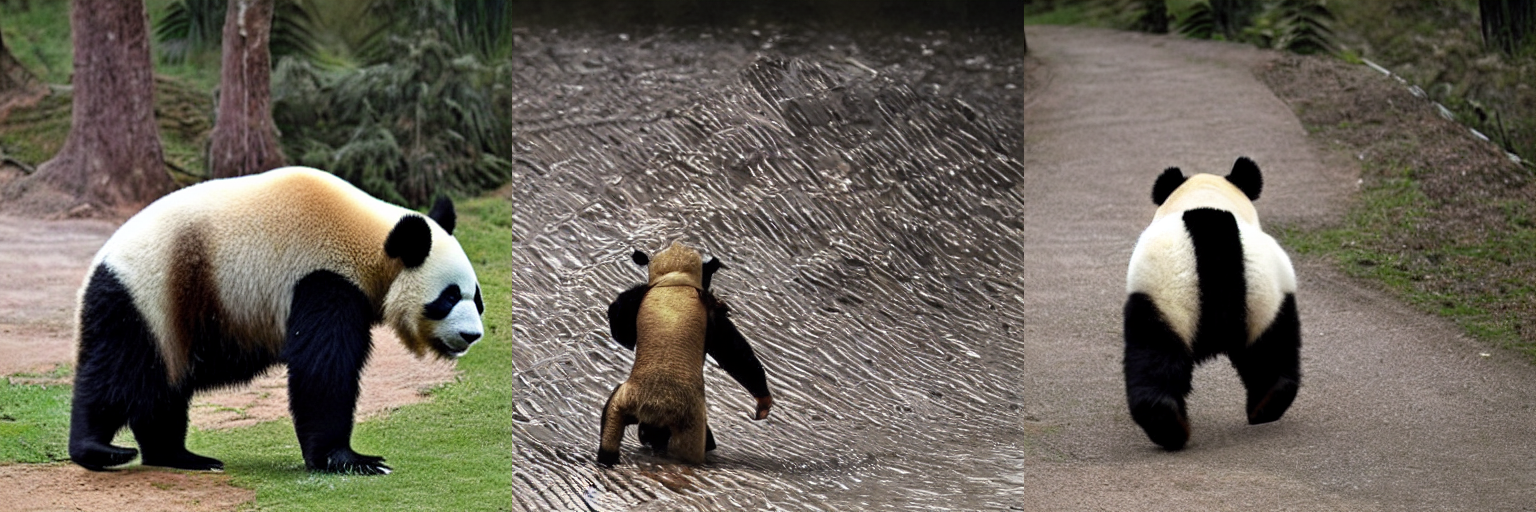}
    \includegraphics[width=0.9\columnwidth]{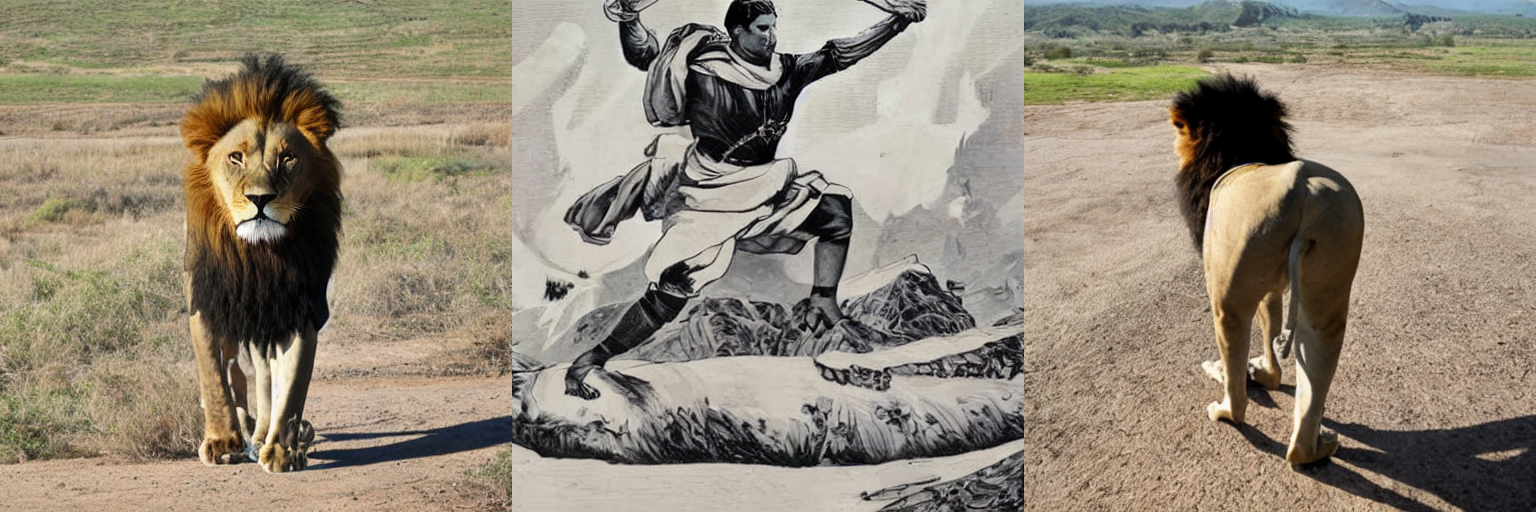}
    \includegraphics[width=0.9\columnwidth]{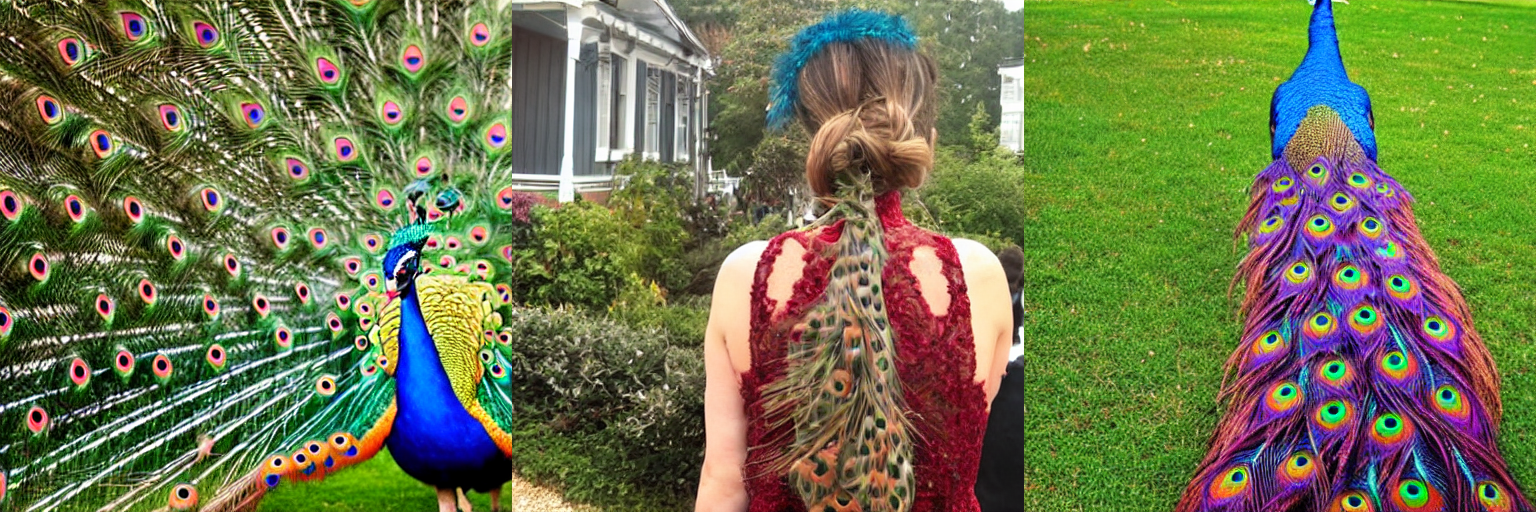}
    \caption{Comparison of generation of the back view of panda, lion and peacock using the vanilla sampler, CEBM, and our Perp-Neg (\textit{from left to right}) with Stable Diffusion.}
    \label{fig:qualitative_comparison_back} 
\end{figure}

\noindent\textbf{Average success rate:} We test each group of prompts using three objects, ``panda", ``lion", and ``peacock" and only count photo-realistic generation that matches the text prompt query as a successful generation. For detailed acceptance criteria, please refer to Appendix~\ref{sec:criteria}. On the side view and back view generation, in each group, we adopt 3 combinations of the complementary view into the negative prompts, \textit{e.g.}, for the case that side view is used in the positive prompt, we use front view, back view and both front and back view in the negative prompt. Then we count the averaged percentage of accepted successful generations, summarized in Table~\ref{tab:sucess_rate}. 

As shown, we can observe the vanilla sampling from Stable Diffusion only has 42.0\% in successfully generating requested side-view images. For more difficult cases like generating the back-view images, the success rate is even lowered to 14.6\%. By simply using negative prompts without considering the overlap between positive and negative prompts, CEBM fails to generate the desired view and results in a lower success rate compared to Stable Diffusion. Compared to these two types of baseline, Perp-Neg shows the effectiveness of properly using negative prompts and significantly improves the success rate by a large margin. \Figref{fig:qualitative_comparison_back} shows a qualitative justification corresponding to Table~\ref{tab:sucess_rate}. From the left column, we can observe without using negative prompts Stable Diffusion may generate incorrect views though requested for the back view. Although using negative prompts, CEBM does not consider the overlap between positive and negative prompts, resulting in artifacts or vanish of the content, shown in the middle column. Different than the previous two failed cases, Perp-Neg is able to properly use negative prompts to eliminate the wrong view and preserve the corresponding details of the text prompt query to achieve realistic generation well-aligned to the input text.

\begin{table}[t]
\centering
\renewcommand{\arraystretch}{1.}
\setlength{\tabcolsep}{1.0mm}{ 
\begin{tabular}{l|c|c}
\toprule[1.5pt]
Method           & Side view & Back view \\ \hline
Stable Diffusion & 42.0\%    & 14.6\%    \\
CEBM             & 12.7\%    & 2.0\%     \\
Perp-Neg (Ours)             & \textbf{73.1}\%    & \textbf{40.4}\%   \\ 
\bottomrule[1.5pt]
    \end{tabular}
    }\vspace{-2mm}
\caption{Comparison of successful generation rate.}%
\label{tab:sucess_rate}
\vspace{-10pt}
\end{table}

\begin{figure}[t]
    \centering
    \includegraphics[width=0.9\columnwidth]{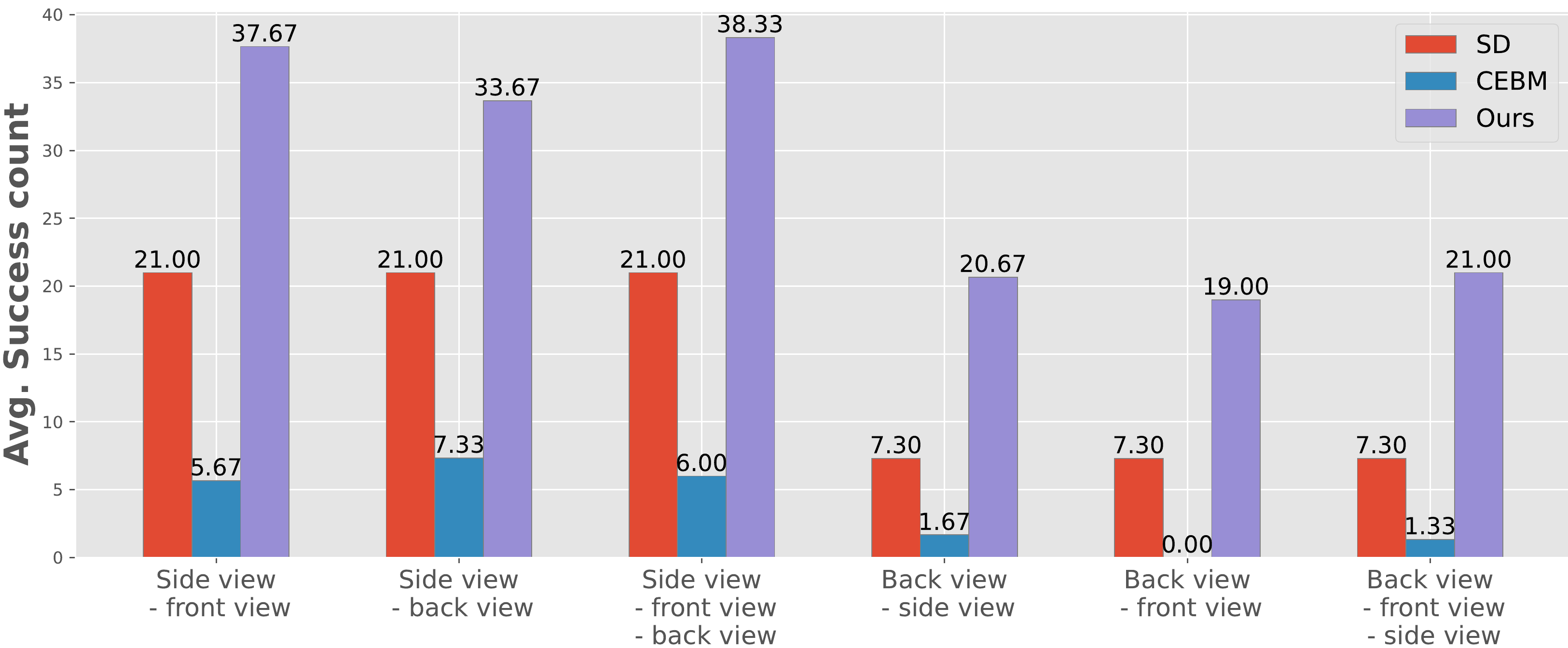}\vspace{-3mm}
    \caption{\small Averaged successful generation count in terms of different positive and negative prompt combinations.}
    \label{fig:case-text} \vspace{-5mm}
\end{figure}

\textbf{On the combination of positive and negative prompts: } To explore how to combine negative prompts with positive prompt. We compute the averaged successful generation count across all tested objects and report the averaged count using different positive and negative prompt combinations in \Figref{fig:case-text}. From the figure, it is notable that when generating the side-view images, using the back view as the negative prompt is less effective than using the front view or using the combination of both the front view and back view. Similarly, when generating back-view images, using the front view in negative prompts is also less effective, since the model is less likely to generate front-view details when conditioned on the back view, while the side view is more ambiguous to the model. This observation indicates putting ambiguous perspectives in the negative prompt could help the model avoid generating undesired images.

\subsection{Perp-Neg DreamFusion}

We integrated Perp-Neg with DreamFusion by using the publicly available replication of DreamFusion that utilizes Stable Diffusion as the pre-trained 2D diffusion model instead of Imagen. We replaced the SDS loss with the one provided in Equation~\ref{eq:sds_loss_perpneg}. To determine the negative prompt weight functions $f$, we used the general form of a shifted exponential decay of the form $f_.(r) = a \exp(-b * r) + c$, where $a$, $b$, and $c$ are greater than or equal to zero. We set the parameters of the $f$ functions separately for each text prompt by generating 2D interpolation samples for 10 different random seeds and then selecting the parameters with the highest accuracy in following the conditioned view. We observed that parameters that allow better interpolation in 2D cases directly relate to the parameters that better help with the Janus problem in 3D. We also noted that if the interpolation between two views remained unchanged while varying the angle for a large range, then the 3D-generated scene was more likely to have a flat geometry from multiple views, resulting in the Janus problem. To overcome this, we perturbed the interpolation factor $r$ with random noise, calculated the interpolated text embedding and their related negative weights, and thus, the model is less likely to generate identical photos from a range of views.

To evaluate the effectiveness of the Perp-Neg in alleviating the Janus problem, we conducted our experiments using prompts that did not depict circular objects. For each prompt, we utilized the Stable DreamFusion method with and without the Perp-Neg algorithm, running 14 trials for each approach with different seeds. Our results indicate the number of successful outputs generated without a Janus problem when using the Perp-Neg algorithm: ``a corgi standing'' 2 times, ``a westie'' 5 times, ``a lion'' 1 time, ``a Lamborghini'' 5 times, ``a cute pig'' 0 times, and ``Super Mario'' 4 times. In contrast, when we ran the model without the Perp-Neg, it failed to generate any correct output except for  ``a Lamborghini'' 4 times and ``Super Mario'' 2 times. These findings clearly demonstrate the advantages of utilizing the Perp-Neg in mitigating the Janus problem. 

\begin{figure}
    \centering
    \includegraphics[width=0.8\columnwidth]{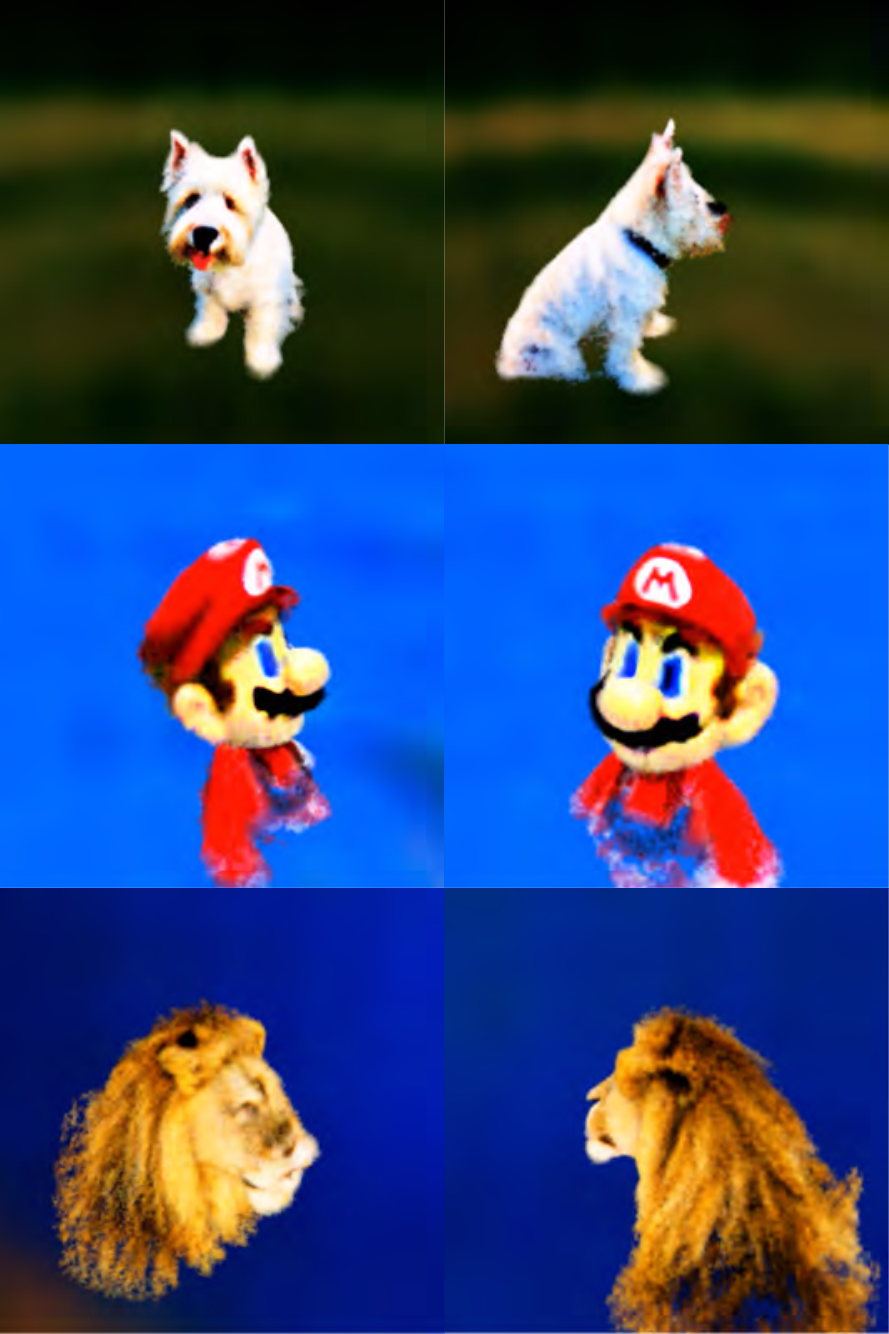}\vspace{-3mm}
    \caption{\small Qualitative examples of Stable Dreamfusion with Perp-Neg, using prompts ``a westie", ``Super Mario" and ``a lion."}
    \label{fig:3d_qualitative}\vspace{-6mm}
\end{figure}

\section{Conclusion}

We introduce %
Perp-Neg, a new algorithm that enables negative prompts to overlap with positive prompts without damaging the main concept.  Perp-Neg provides greater flexibility in generating images by enabling users to edit out unwanted concepts from initial generated photos. More importantly, Perp-Neg enhances prompt faithfulness by preventing the 2D diffusion model from producing biased samples from its training data and accurately representing the input prompt. This can be accomplished by feeding to Perp-Neg a sentence describing the model bias as the negative prompt to generate desired solutions. 
Our paper also demonstrates how Perp-Neg can properly condition the 2D diffusion model to generate views of interest rather than a canonical view. Finally, we integrate Perp-Neg's robust view conditioning property into SDS-based text to 3D models and show how it alleviates the Janus problem.

{\small
\bibliographystyle{ieee_fullname}
\bibliography{ref}
}

\clearpage
\appendix

\section*{Supplementary Materials}
\section{Experiment details}

\subsection{Implementation details}
\begin{figure*}[h]
    \centering
    \includegraphics[width=\textwidth]{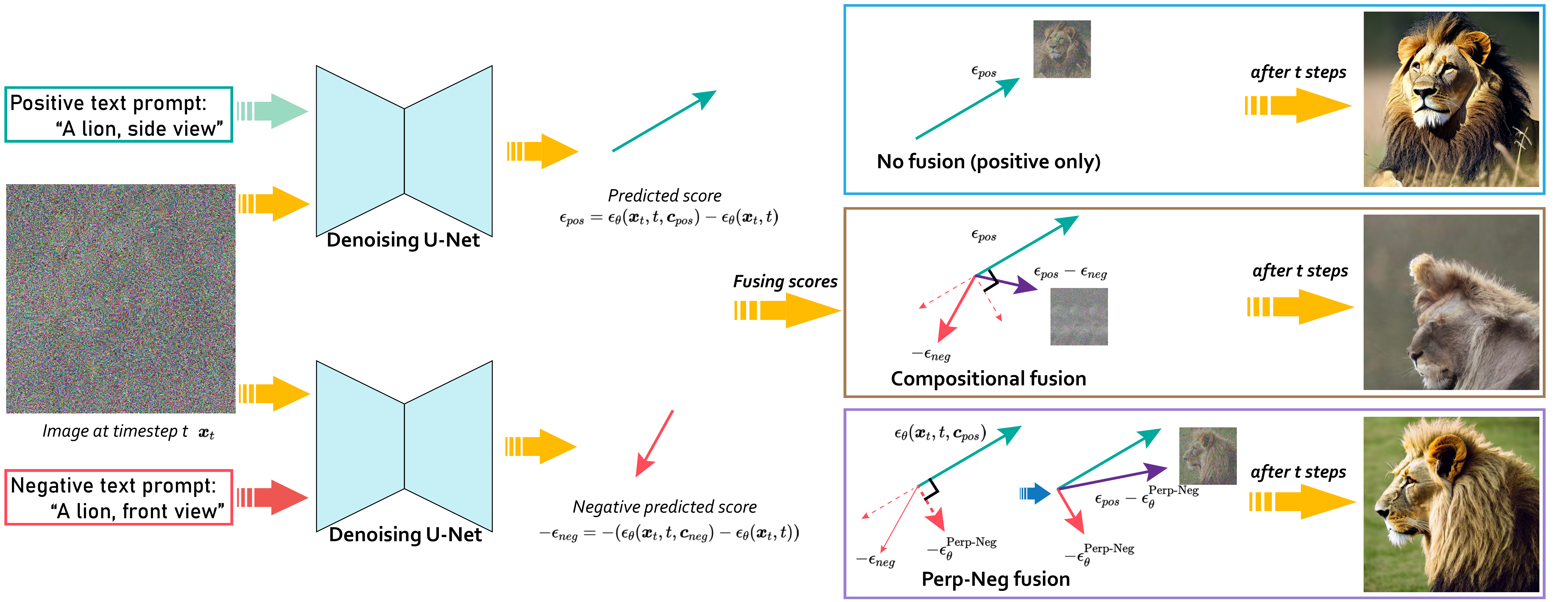}\vspace{-2mm}
    \caption{Illustrative depiction of Perp-Neg, and comparison with sampling without the usage of negative prompt and compositional negative prompt fusion. }
    \label{fig:model}
\end{figure*}

To clarify the difference between with/without negative prompts, and with/without Perp-Neg, we provide an illustrative depiction in \Figref{fig:model}. Our Perp-Neg does not require additional training or fine-tuning, and is implemented in the sampling pipeline. A detailed implementation in each timestep is shown in Algorithm~\ref{alg:code}. 
 
\begin{algorithm}[h]
\centering
\caption{Perp-Neg Pseudocode at timestep $t$, PyTorch-like style}
\label{alg:code}
\definecolor{codeblue}{rgb}{0.25,0.5,0.5}
\definecolor{codekw}{rgb}{0.85, 0.18, 0.50}
\lstset{
  backgroundcolor=\color{white},
  basicstyle=\fontsize{7.5pt}{7.5pt}\ttfamily\selectfont,
  columns=fullflexible,
  breaklines=true,
  captionpos=b,
  commentstyle=\fontsize{7.5pt}{7.5pt}\color{codeblue},
  keywordstyle=\fontsize{7.5pt}{7.5pt}\color{codekw},
}
\begin{lstlisting}[language=python]
# epsilon: diffusion model 
# x: the noisy image at the current timestep
# t: the current time step t
# c: main text prompt
# c_neg: auxiliary negative text prompt list
# a: classifier-free guidance scale
# w_pos: weight of main text prompts
# w_neg: weight list of negative text prompts

def Perp_neg(M, x, t, c, c_neg, a, w_pos, w_neg):
    # get the main component
    e_main = epsilon(x, t, c) - epsilon(x, t) 
    
    # proceed with auxiliary negative text prompts
    for i, text_negative in enumerate(c_neg):
        e_i = epsilon(x, t, text_negative) - epsilon(x, t)
        accum_grad -=  w_neg[i] * get_prependicualr_component(e_i, e_main) # accumulate the negative gradients in the opposite direction
    
    e = epsilon(x, t) +  a * (w_pos * z_main + accum_grad) # update the noisy image at the current timestep to the next step (eq. 8)
    
    return e # return the final prediction at time step t.

    
# definition of "get_prependicualr_component"
def get_prependicualr_component(x, y):
    # x: gradient of principle component
    # y: gradient of auxiliary component
    proj_x = ((torch.mul(x, y).sum())/(torch.norm(y)**2)) * y  # cosine projection of x on y
    return x - proj_x # get perpendicular vector of x

\end{lstlisting}
\end{algorithm}

For 2D generation, we implement our Perp-Neg into Stable Diffusion v1.4 pipeline. We adopt 50 DDIM steps, and fix the guidance scale as $a=7.5$ and the positive weight $w_\text{pos}=1$, for each generation. For negative weight, the value may vary and we find generally in the range [-5, -0.5] can produce satisfactory images. In the 2D view generation experiments, we fix the negative weights to $w_\text{neg}=-1.5$ when there is only one negative prompt, and set $w_{\text{neg}_1} \!=\! w_{\text{neg}_2} \!=\! -1$ when there are two negative prompts. The results are generated across seeds 0-49. For the interpolation between two views, we normally interpolate $r_{inter}$ with stride 0.25 between 0 and 1 to have 5 images in total. Moreover, for 3D generation, we employ Perp-Neg into Stable Dreamfusion. The results of our baselines are reproduced with their open-source code\footnote{https://github.com/energy-based-model/Compositional-Visual-Generation-with-Composable-Diffusion-Models-PyTorch}\footnote{https://github.com/ashawkey/stable-dreamfusion}. 
All experiments are conducted using Pytorch 1.10 on a single NVIDIA-A5000 GPU. 

\subsection{Criteria for successful view generation count}\label{sec:criteria}
\begin{figure*}[t]
    \centering
    \includegraphics[width=\textwidth]{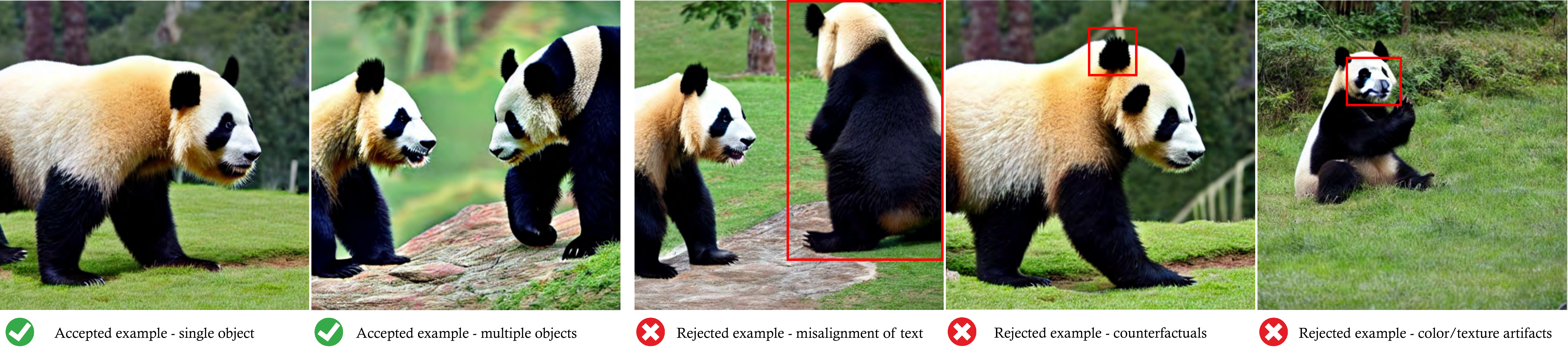}\vspace{-3mm}
    \caption{Example of the accepted/rejected generation.}\vspace{-4mm}
    \label{fig:criteria} 
\end{figure*}

Here we elaborate on the criteria for the successful generation count in our quantitative experiments in Section~\ref{sec:quantitative exp}. We reject the image samples with the following criteria and examples are shown in \Figref{fig:criteria}:
\begin{itemize}
    \item The images that do not show requested object(s) or view. Note that if the generated image contains multiple objects, and one of them is not positioned in the correct view, the image will still be rejected.
    \item The images show hallucination including counterfactual details, for example, a panda has three ears.
    \item The images have color or texture artifacts that make the images not realistic. 
\end{itemize}

\section{Additional experiments}

\subsection{Case-by-case study}
We further conduct case-by-case studies for the previous experiments, where we collect statistics of every tested object per view. We report the averaged acceptance rate across all possible positive and negative combinations in Table~\ref{tab:case-obj}. 

From the table, we find back view is consistently more difficult to generate than the side view. A possible explanation is that the generator may have more information about the front view from the training datasets, and the side view possesses more connection to the front view, while it requires additional knowledge to generate the corresponding back view. In terms of the objects, we find it is less likely to generate faithful images of peacock(s) in the side view, as well as peacock(s) in the back view lion in the back view, as there may have less corresponding training images in the original training datasets.

\begin{table}[h]
\vspace{0mm}
\centering
\begin{adjustbox}{width=\columnwidth}
\renewcommand{\arraystretch}{1.}
\setlength{\tabcolsep}{1.0mm}{ 
\begin{tabular}{l|ccc|ccc}
\toprule[1.5pt]
View   & \multicolumn{3}{c|}{Side}  & \multicolumn{3}{c}{Back}  \\ \hline
Object & Lion   & Panda  & Peacock & Lion   & Panda  & Peacock \\ \hline
SD     & 58.0\% & 44.0\% & 24.0\%  & 8.0\%  & 28.0\% & 8.0\%   \\
CEBM   & 14.0\% & 13.3\% & 10.7\%  & 0\%    & 4.0\%  & 2.0\%   \\
Ours   & \textbf{80.7}\% & \textbf{83.3}\% & \textbf{55.3}\%  & \textbf{49.3}\% & \textbf{34.0}\% & \textbf{38.0}\%    \\ 
\bottomrule[1.5pt]
\end{tabular}
}
\end{adjustbox}
\caption{Case-by-case percentage of successful view generations different objects.}
\label{tab:case-obj}
\end{table}

\subsection{Ablation study}

\begin{figure*}[h]
    \centering
    \includegraphics[width=\textwidth]{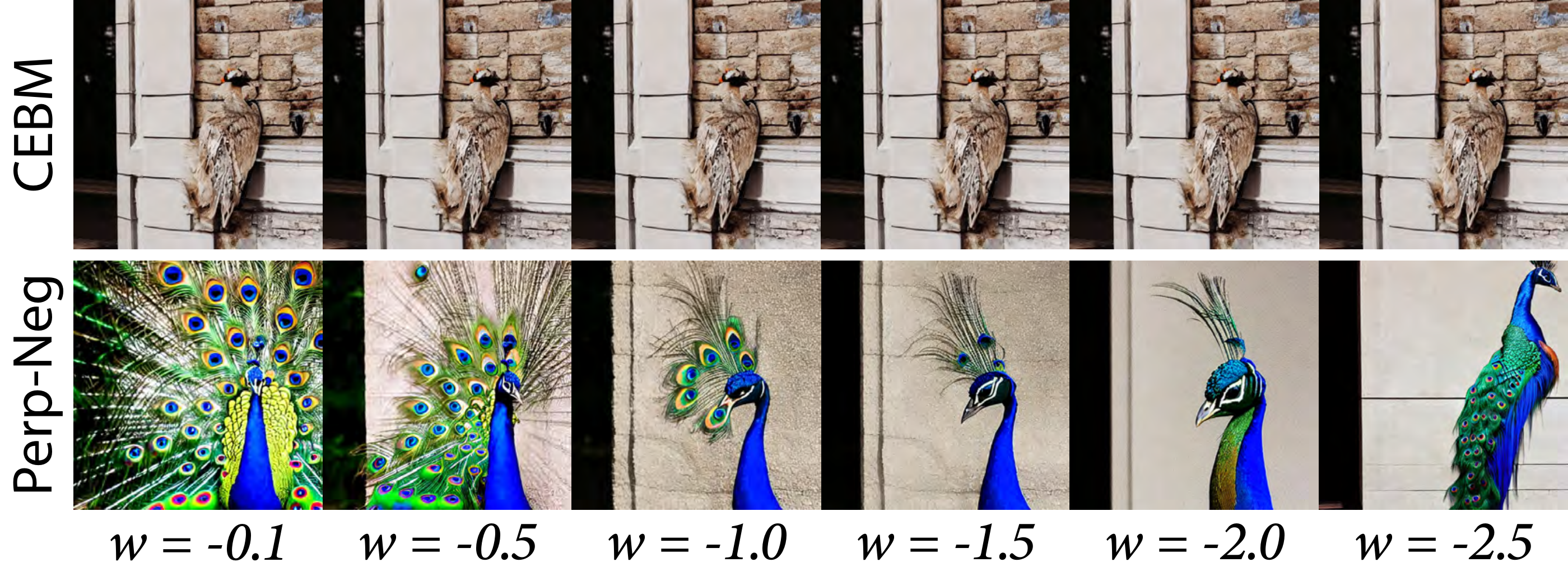}
    \caption{Ablation studies with different negative prompt weights in the generation, side view of peacocks.}
    \label{fig:ablation_peacock} 
\end{figure*}

\begin{figure*}[h]
    \centering
    \includegraphics[width=\textwidth]{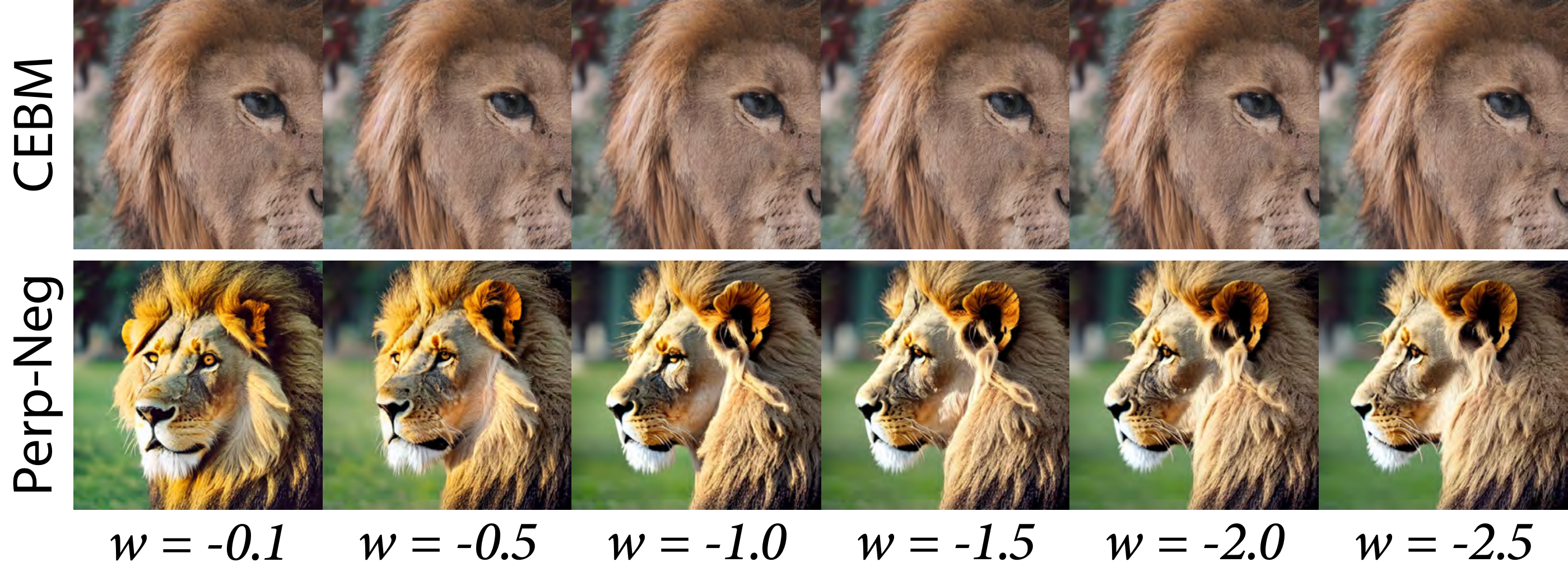}
    \caption{Analogous visualization of lions to \Figref{fig:ablation_peacock}.}
    \label{fig:ablation_lion} 
\end{figure*}

\begin{figure*}[h]
    \centering
    \includegraphics[width=\textwidth]{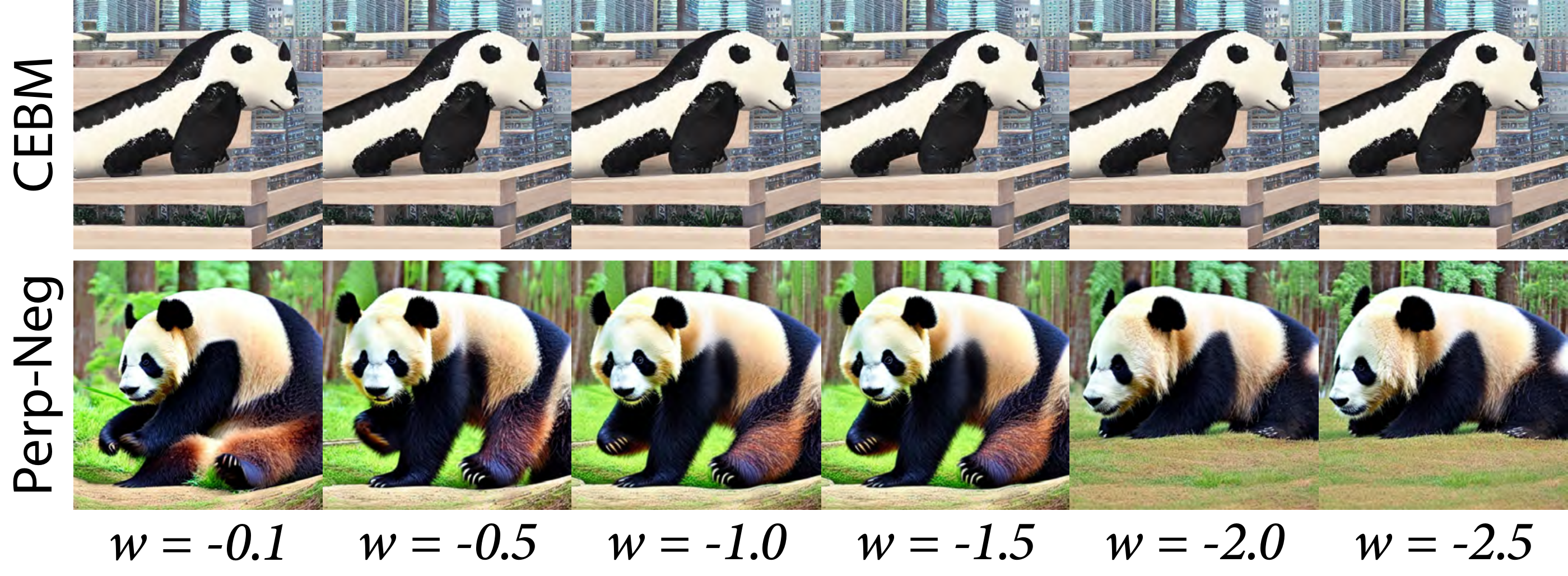}
    \caption{Analogous visualization of pandas to \Figref{fig:ablation_peacock}.}
    \label{fig:ablation_panda} 
\end{figure*}

We also provide ablation studies on the effects of negative prompt weights to see how the weight affects the usage of negative attribute elimination. We show the results of generations using different negative prompt weights $w_{i}$ in \Figref{fig:ablation_peacock}-\ref{fig:ablation_panda}. We can observe for CEBM, the results are consistently not relevant to the requested text content, no matter the negative prompt weights are small or big. For Perp-Neg, we can see with larger weights, \textit{e.g.}, $w=-0.1$, the generated results are not positioned in the side view. As we decrease the weight, the generated image becomes more relevant to the text. This observation indicates Perp-Neg has better controllability in eliminating the negative attributes in the prompts.  

\subsection{Additional results}
In the following, we provide additional qualitative results of 2D generation and view interpolation. And for 3D generation results, please refer to the provided video in the supplementary file. 
\begin{figure*}[h]
    \centering
    \includegraphics[width=\textwidth]{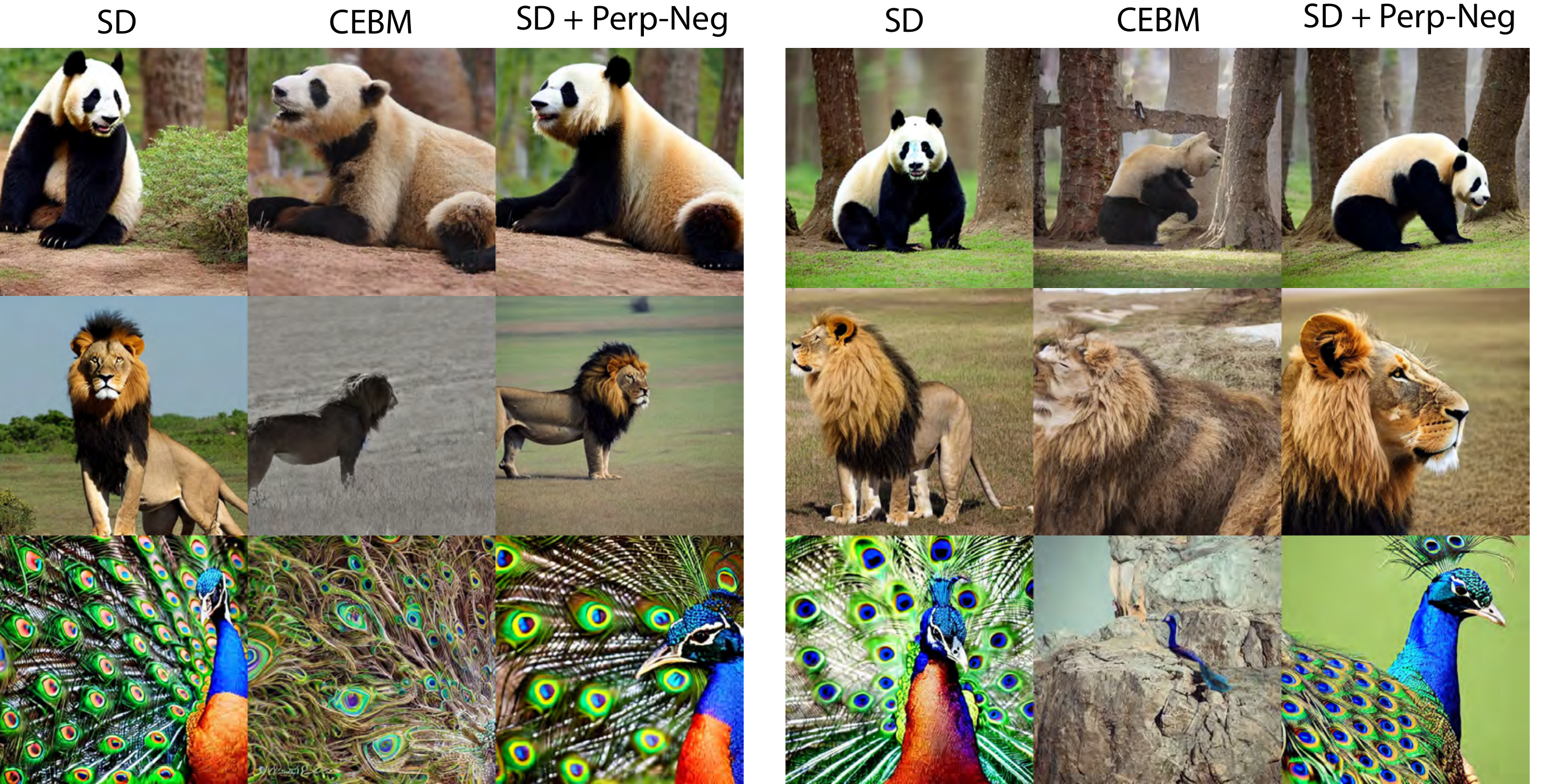}
    \caption{Analogous visualization to \Figref{fig:qualitative_comparison_back},  side-view generation.}
    \label{fig:qualitative_comparison_side} 
\end{figure*}

\begin{figure*}[h]
    \centering
    \includegraphics[width=\textwidth]{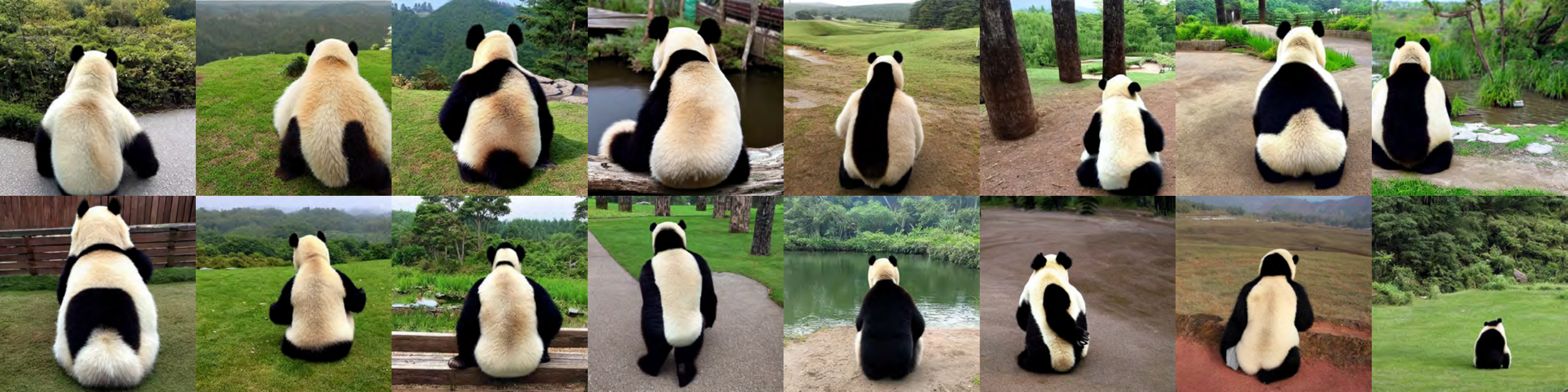}
    \caption{Additional visualization of panda back-view from our experiment. Here we provide a part of those generations with seeds 0-49 using Perp-Neg, including both successful and failed samples. Most of the generations show the semantics of ``panda back view". }
    \label{fig:qualitative_panda_back} 
\end{figure*}

\begin{figure*}[h]
    \centering
    \includegraphics[width=\textwidth]{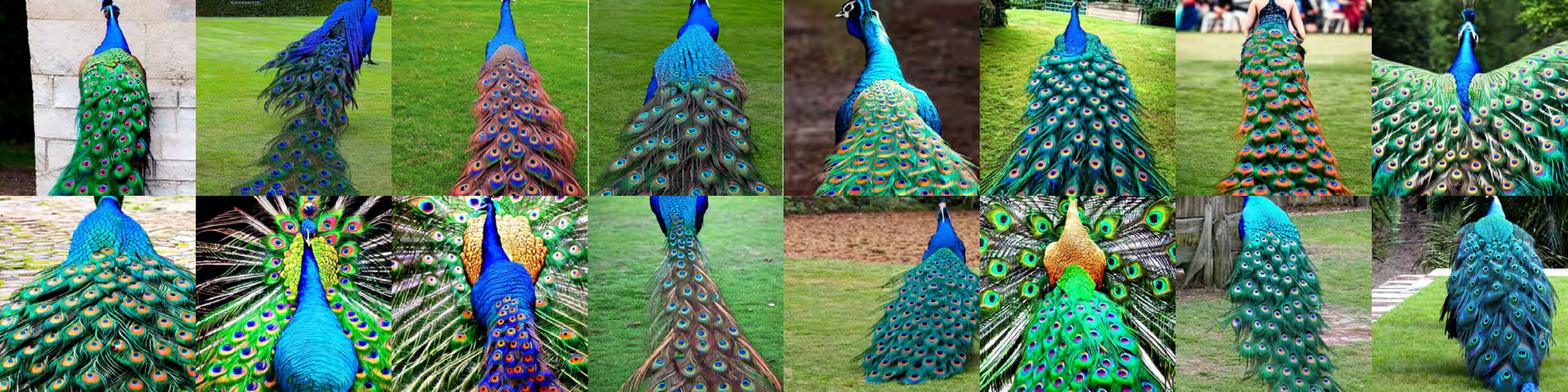}
    \caption{Analogous visualization to \Figref{fig:qualitative_panda_back} visualization of peacock back-view.}
\end{figure*}

\begin{figure*}[h]
    \centering
    \includegraphics[width=\textwidth]{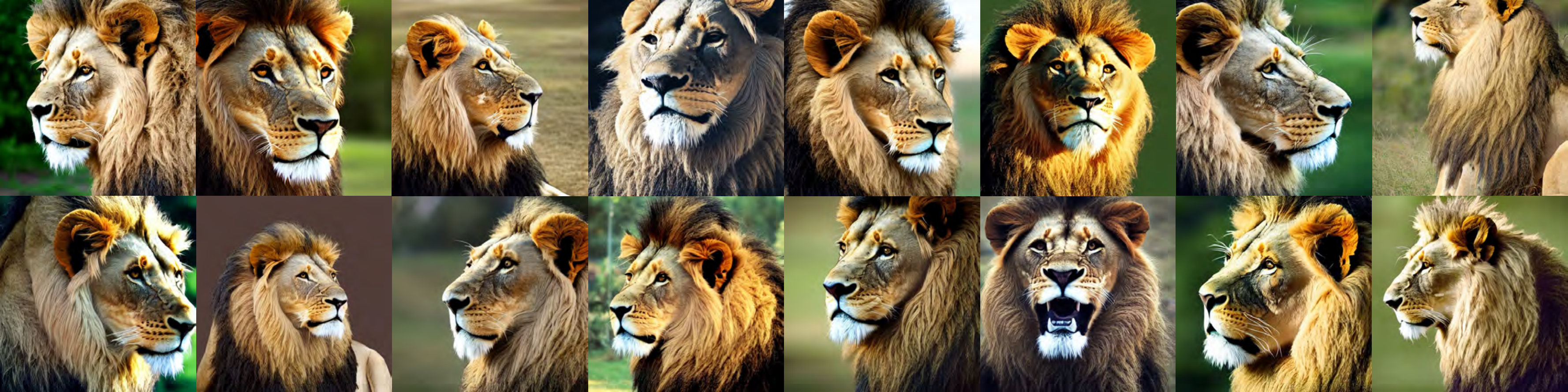}
    \caption{Analogous visualization to \Figref{fig:qualitative_panda_back} visualization of lion side-view.}
    \label{fig:qualitative_lion_side} 
\end{figure*}

\begin{figure*}[h]
    \centering
    \includegraphics[width=\textwidth]{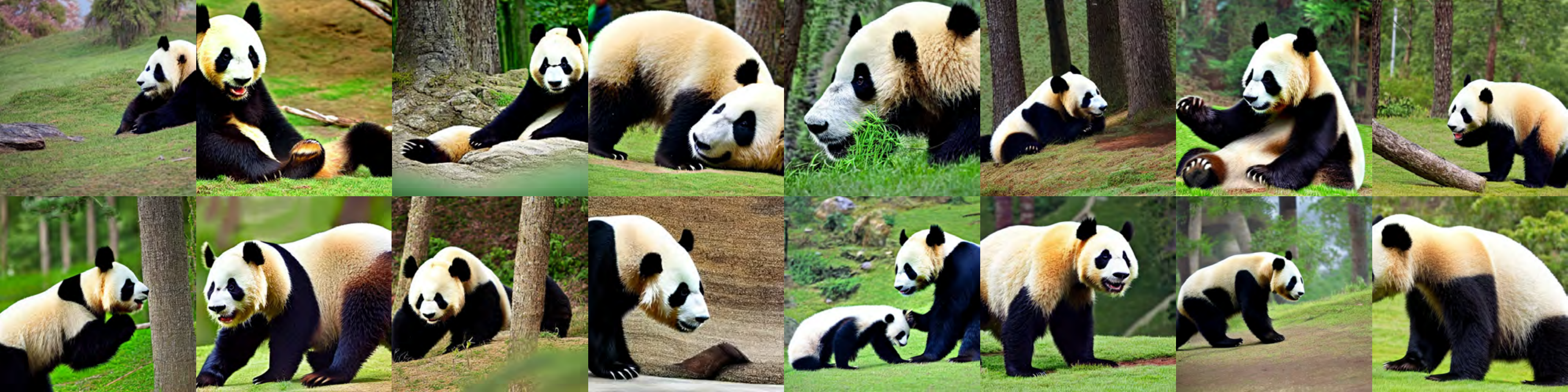}
    \caption{Analogous visualization to \Figref{fig:qualitative_panda_back} visualization of panda side-view.}
    \label{fig:qualitative_panda_side} 
\end{figure*}

\begin{figure*}[h]
    \centering
    \includegraphics[width=\textwidth]{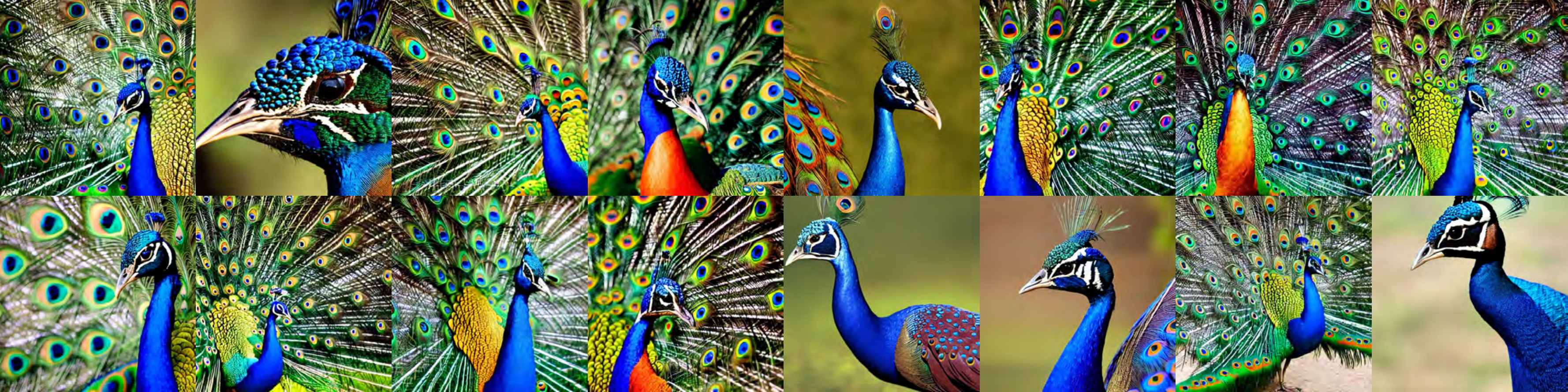}
    \caption{Analogous visualization to \Figref{fig:qualitative_panda_back} visualization of peacock side-view.}
    \label{fig:qualitative_peacock_side} 
\end{figure*}

\begin{figure*}[ht]
  \centering
\includegraphics[width=0.9\textwidth]{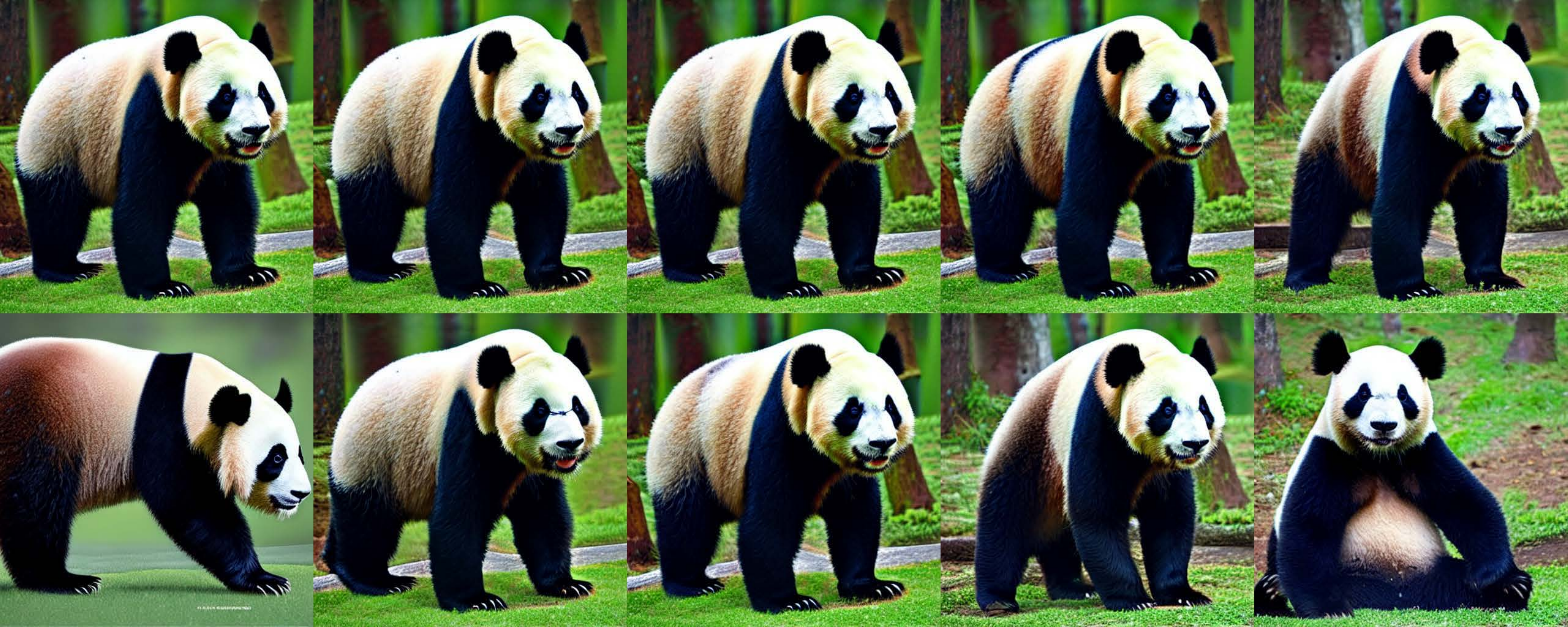}
  \vspace{-10pt}
      \[
    \text{a giant panda, side view} \xrightarrow{\quad\quad} \text{a giant panda, front view}
    \]
  \vspace{-10pt}

  \includegraphics[width=0.9\textwidth]{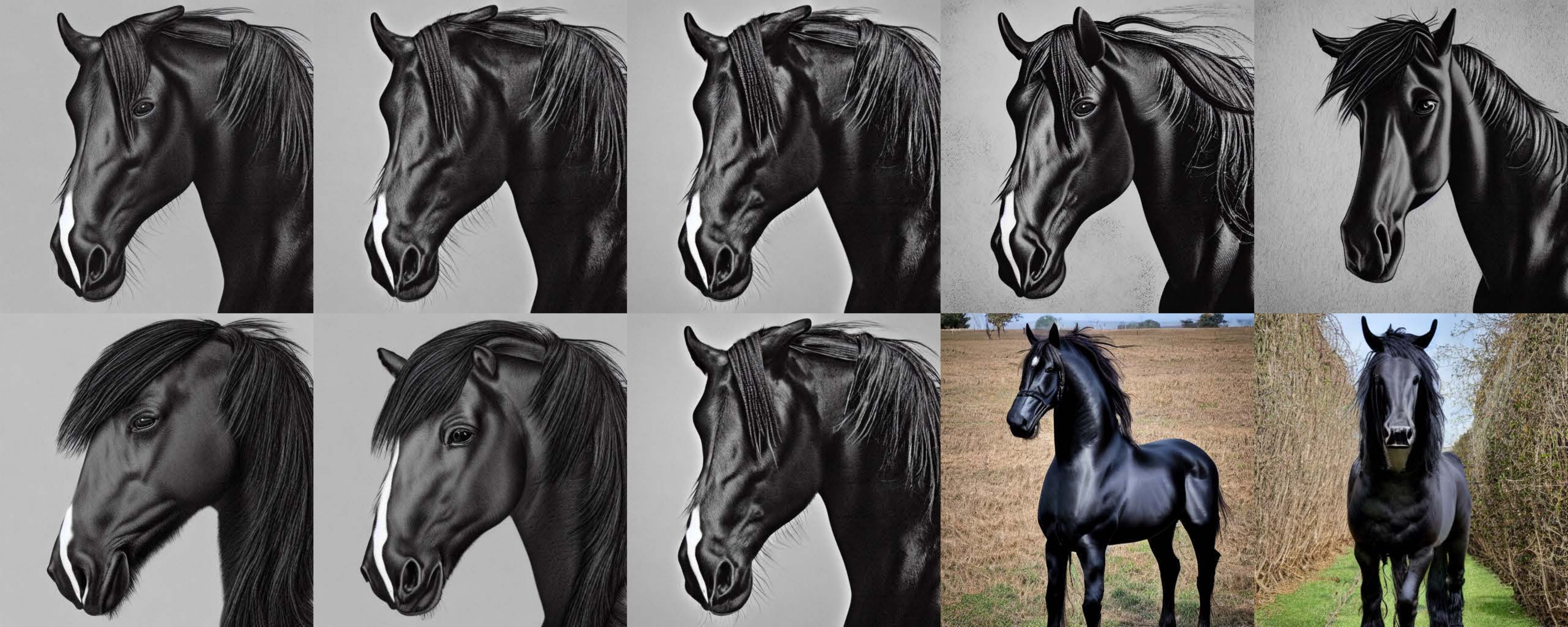} %
  \vspace{-10pt}
      \[
    \text{a Friesian horse, side view} \xrightarrow{\quad\quad} \text{a a Friesian horse, front view}
    \]  
\includegraphics[width=0.9\textwidth]{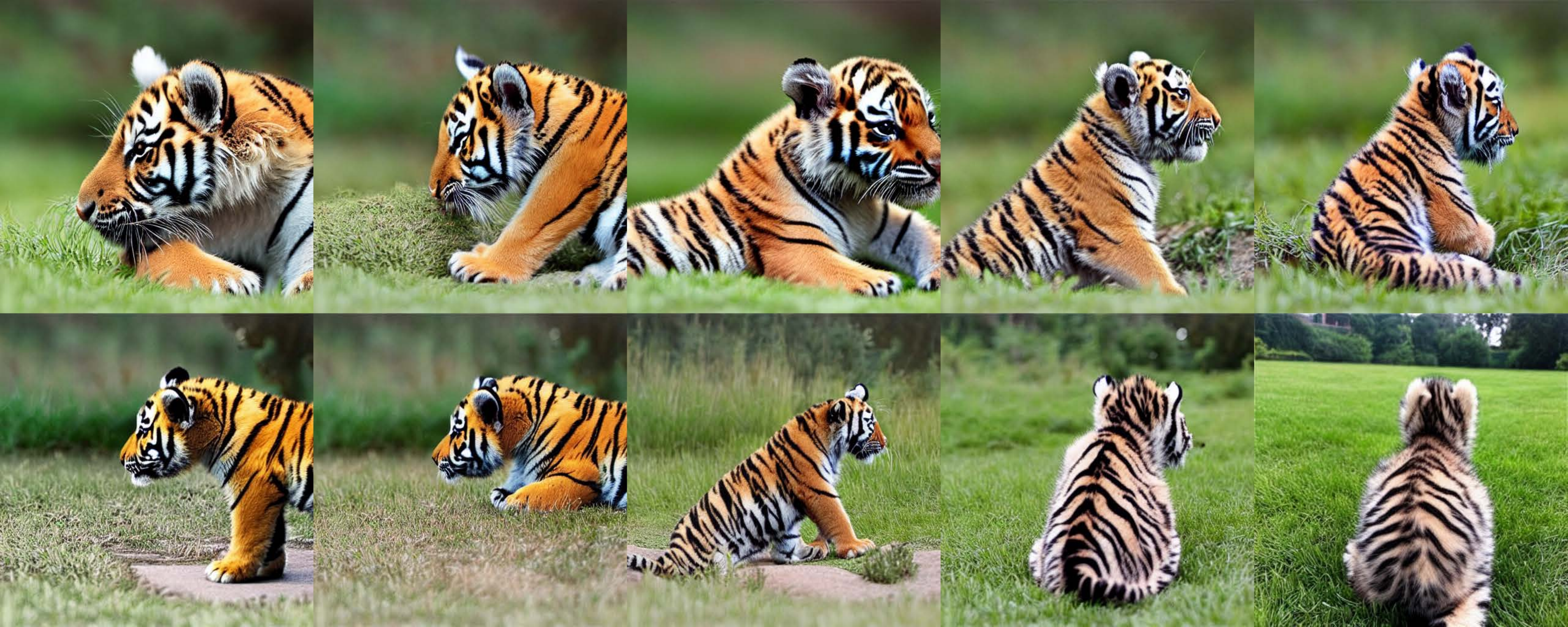} %
  \vspace{-10pt}
      \[
    \text{a cute tiger cub, side view} \xrightarrow{\quad\quad} \text{a cute tiger cub, back view}
    \]    
  \vspace{-20pt}
   {\caption*{}}
\end{figure*}
\begin{figure*}[ht]
  \centering
  \vspace{-150pt}
\includegraphics[width=0.9\textwidth]{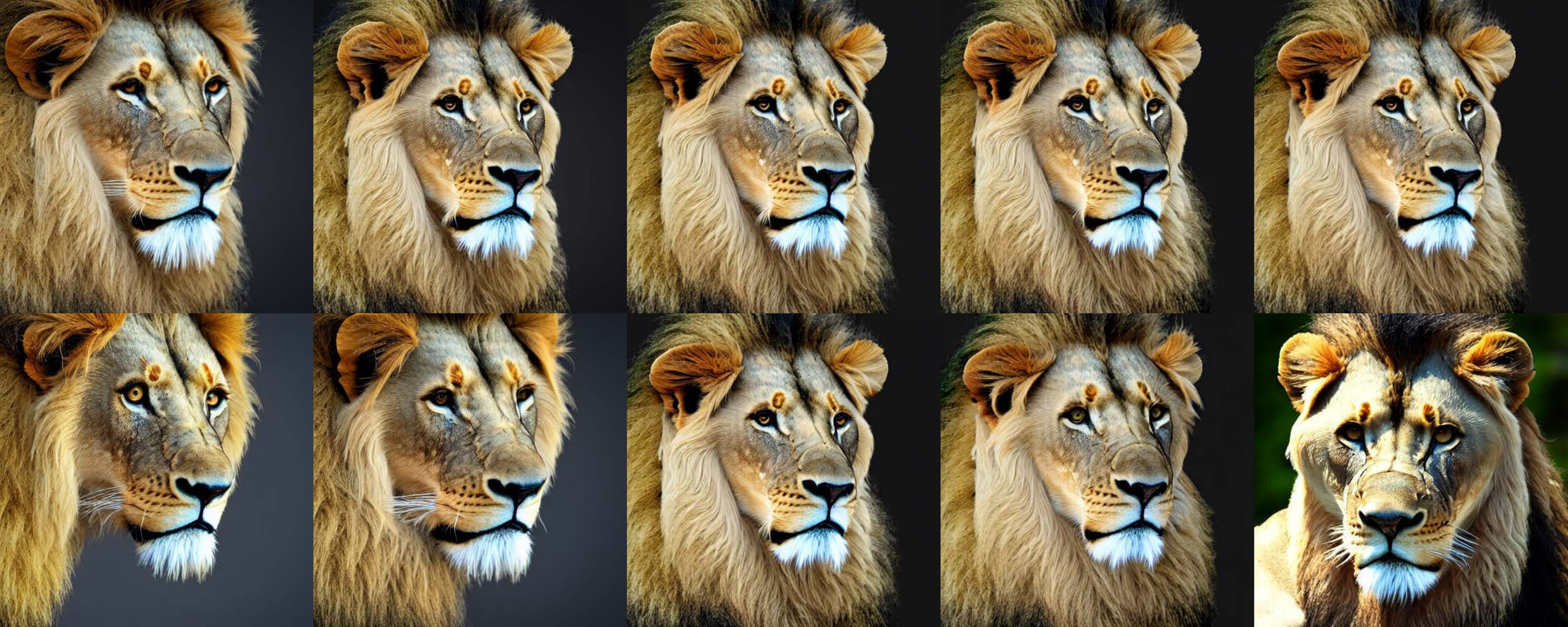}
  \vspace{-10pt}
      \[
    \text{a lion, side view} \xrightarrow{\quad\quad} \text{a lion, front view}
    \]
  \vspace{-10pt}

  \includegraphics[width=0.9\textwidth]{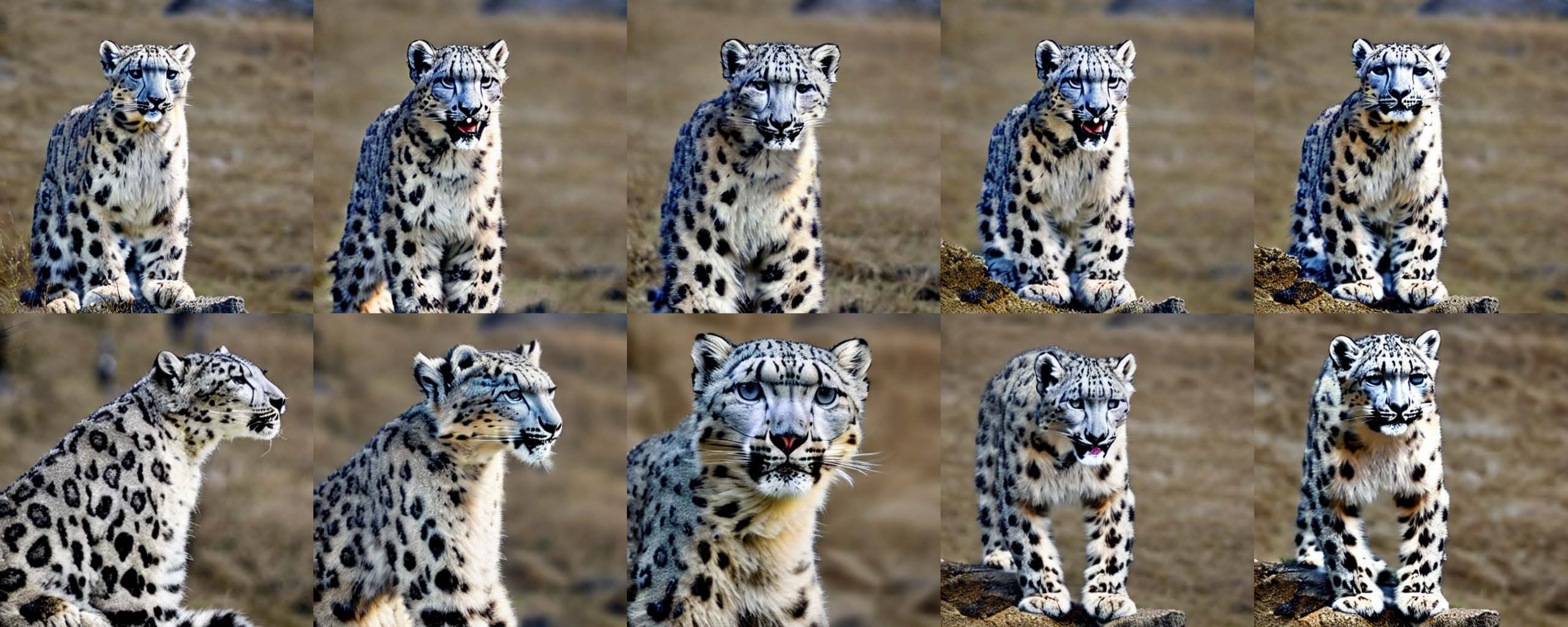} %
  \vspace{-10pt}
      \[
    \text{a snow leopard, side view} \xrightarrow{\quad\quad} \text{a snow leopard, front view}
    \]  
  \vspace{-20pt}
  \caption{\textbf{Qualitative comparison} of view interpolation with/without Perp-Neg. We fixed the seed across different images of each prompt. For each prompt, the top row shows the result of text-embedding interpolation without Perp-Neg. And, the bottom row shows the result of Perp-Neg.}
  \label{fig:fullwidthphoto_tiger_}
\end{figure*}
\clearpage

\end{document}